\documentclass[10pt,twocolumn,letterpaper]{article}

\usepackage[accsupp]{axessibility}  

\usepackage[pagenumbers]{cvpr} 

\definecolor{cvprblue}{rgb}{0.21,0.49,0.74}
\usepackage[pagebackref,breaklinks,colorlinks,allcolors=cvprblue]{hyperref}

\usepackage{multirow}
\usepackage{adjustbox}
\usepackage{arydshln}
\usepackage{amssymb}
\usepackage{pifont}
\newcommand{\cmark}{\ding{51}}%
\def\paperID{2621} 
\def\confName{CVPR}
\def\confYear{2026}

\title{GeoMMBench and GeoMMAgent: Toward Expert-Level\\Multimodal Intelligence in Geoscience and Remote Sensing\vspace{-10pt}}

\author{Aoran Xiao$^{*,1,\dagger}$, Shihao Cheng$^{*,2}$, Yonghao Xu$^{3}$, Yexian Ren$^{5}$, 
Hongruixuan Chen$^{1,4}$,
Naoto Yokoya$^{1,4}$\\
$^{1}$RIKEN AIP 
\qquad
$^{2}$Wuhan University 
\qquad 
$^{3}$Linköping University
\qquad
$^{4}$University of Tokyo\\
$^{5}$Nanjing University of Information Science and Technology\\  
\vspace{-15pt}
}

\begin{document}
\maketitle
\renewcommand{\thefootnote}{\fnsymbol{footnote}}
\begin{abstract}
\footnotetext{
$^*$ Equal contribution; 
$^\dagger$ corresponding author}
Recent advances in multimodal large language models (MLLMs) have accelerated progress in domain-oriented AI, yet their development in geoscience and remote sensing (RS) remains constrained by distinctive challenges: wide-ranging disciplinary knowledge, heterogeneous sensor modalities, and a fragmented spectrum of tasks. To bridge these gaps, we introduce GeoMMBench, a comprehensive multimodal question-answering benchmark covering diverse RS disciplines, sensors, and tasks, enabling broader and more rigorous evaluation than prior benchmarks. Using GeoMMBench, we assess 36 open-source and proprietary large language models, uncovering systematic deficiencies in domain knowledge, perceptual grounding, and reasoning—capabilities essential for expert-level geospatial interpretation. Beyond evaluation, we propose GeoMMAgent, a multi-agent framework that strategically integrates retrieval, perception, and reasoning through domain-specific RS models and tools. Extensive experimental results demonstrate that GeoMMAgent significantly outperforms standalone LLMs, underscoring the importance of tool-augmented agents for dynamically tackling complex geoscience and RS challenges.
Project: \url{https://geo-mm-agi.github.io}
\end{abstract}    
\section{Introduction}
\label{sec:intro}

Recent progress in multimodal large language models (MLLMs) has accelerated the pursuit of artificial general intelligence (AGI), marking an important milestone with the emergence of Expert AGI. This level of machine intelligence is characterized by performance comparable to or exceeding that of skilled human experts at the 90th percentile across professional domains~\cite{morris2023levels}. As research advances toward this stage, a growing collection of multimodal benchmarks has been introduced to assess reasoning, perception, and domain expertise across diverse disciplines~\cite{yue2024mmmu,xu2024lvlm,li2023seed,liu2024mmbench,hu2024omnimedvqa}. These benchmarks serve as essential instruments for measuring progress and guiding the development of more capable multimodal systems.

Despite these advances, the domain of geoscience and remote sensing (RS) has received limited attention in current benchmarking efforts. Existing studies~\cite{soni2024earthdial,danish2024geobench,wang2024earthvqa,kuckreja2024geochat,irvin2024teochat} remain narrow in scope, focusing primarily on perception tasks such as classification, detection, or segmentation, and are often restricted to a limited set of sensor modalities, particularly optical imagery. In contrast, geoscience requires the integration of multi-sensor data, spatiotemporal reasoning, and multidisciplinary knowledge, which existing benchmarks fail to adequately evaluate. This gap highlights the need for a comprehensive and expert-level assessment framework to systematically measure and advance AGI performance within the RS domain.

To bridge this gap, we introduce \textbf{GeoMMBench}, a comprehensive domain-specific benchmark designed to evaluate the multidisciplinary and multimodal capabilities of MLLMs in geoscience and RS. The benchmark contains 1,053 expert-level, image-based multiple-choice questions that represent diverse and realistic challenges across the key dimensions, as illustrated in Fig.~\ref{fig1:dimensions}.

GeoMMBench is characterized by three major properties that define its breadth and depth. (1) \textit{Multi-disciplinary coverage}: It spans multiple geospatial disciplines, including RS, photogrammetry, geographic information systems (GIS), and global navigation satellite systems (GNSS). These areas draw upon foundational knowledge in mathematics, physics, cartography, spectroscopy, and geography, mirroring the diverse expertise required for professional interpretation and decision-making. (2) \textit{Multi-sensor modalities}: The benchmark incorporates a wide range of sensing sources such as optical imagery, synthetic aperture radar (SAR), multispectral and hyperspectral data, LiDAR, and digital elevation models (DEM). These complementary modalities enable a realistic evaluation of multi-sensor integration skills crucial for expert-level analysis. (3) \textit{Diverse task spectrum}: GeoMMBench covers a continuum of complexity, from theoretical and data-preprocessing concepts to perception tasks and high-level geospatial applications, providing a holistic assessment of AGI capabilities in RS.

We conduct a comprehensive evaluation of 36 open-source and proprietary MLLMs on GeoMMBench, revealing critical gaps in their ability to interpret geoscience data. These limitations stem from the benchmark’s multidimensional structure, high domain complexity, and requirement for expert-level reasoning. To address these challenges, we propose \textbf{GeoMMAgent}, a multi-agent system that advances MLLMs toward expert-level performance in geoscience and RS. GeoMMAgent integrates large language models (LLMs) with domain-specific tools and models within an adaptive framework that plans, coordinates, and executes tasks across three key functional stages: retrieval, perception, and reasoning. Through strategic task decomposition and tool selection, GeoMMAgent enables models to access external knowledge bases, analyze complex RS data, and perform spatial reasoning beyond the intrinsic capacity of general-purpose MLLMs. The framework supports transparent and modular integration of various RS tools and model components, promoting interpretability, scalability, and future extensibility.

Our main contributions are summarized as follows. \textit{First}, we propose GeoMMBench, a comprehensive and domain-specific benchmark, which evaluates the multidisciplinary, multimodal, and multi-level reasoning capabilities of MLLMs in geoscience and RS.
\textit{Second}, we develop GeoMMAgent, a multi-agent system tailored to geoscience and RS. It integrates MLLMs with domain-specific models and tools through structured task planning that spans retrieval, perception, and reasoning stages, enabling adaptive coordination and expert-level task execution.
\textit{Third}, we conduct an extensive evaluation of open-source and proprietary MLLMs on GeoMMBench, establishing the current performance landscape, identifying key challenges in domain expertise, and demonstrating the effectiveness of GeoMMAgent in bridging these capability gaps.
\section{Related Works}
\label{sec:related_works}

\textbf{MLLMs for Geoscience and Remote Sensing.}
Despite rapid progress in MLLMs, their applications in geoscience and RS remain limited \cite{xiao2025foundation,wei2026sarlang,zhan2025skyeyegpt}. Most recent RS studies focus on VQA over optical RGB imagery~\cite{Zhu_2025_CVPR,Luo_2025_ICCV}. GeoChat~\cite{kuckreja2024geochat} fine-tunes LLaVA-1.5~\cite{liu2024improvedbaselinesvisualinstruction} with detection-based instruction data, while TeoChat~\cite{irvin2024teochat} adapts temporal Earth observation data for change detection. LHRS-Bot~\cite{muhtar2024lhrs} and EarthVQA~\cite{wang2024earthvqa} further tailor multimodal datasets for RS perception and pixel-level understanding. However, these models remain task-specific and lack multi-disciplinary knowledge integration, multi-sensor awareness, and advanced reasoning, which are key capabilities for expert-level intelligence in geoscience and RS.

\noindent\textbf{Knowledge-driven Benchmarks for RS.}
Progress toward Expert AGI~\cite{morris2023levels} has inspired knowledge-oriented benchmarks to assess professional expertise. Early textual evaluations such as AGIEval~\cite{zhong2023agieval}, ARC~\cite{boratko2018systematic}, and MMLU~\cite{hendrycks2020measuring} focus on standardized exams, while MMLU-Pro~\cite{wang2024mmlu} introduces higher-order reasoning. Multimodal extensions like ScienceQA~\cite{lu2022learn}, MMMU~\cite{yue2024mmmu}, MMMU-Pro~\cite{yue2024mmmu_pro}, and Video-MMMU~\cite{hu2025video} broaden evaluation across images and temporal sequences. In contrast, RS benchmarks still concentrate on perception tasks such as scene classification~\cite{wang2024skyscript,lu2017exploring,yuan2022exploring,cheng2022nwpu}, object detection~\cite{pang2024vhmversatilehonestvision,livrsbench}, and semantic segmentation~\cite{wang2024earthvqa,livrsbench}, often using automatically generated descriptions instead of expert assessments. To the best of our knowledge, GeoMMBench is the first expert-crafted, knowledge-driven benchmark that spans multiple disciplines, sensor modalities, and task hierarchies, bridging theoretical concepts and practical applications in geoscience and RS.
\section{GeoMMBench Benchmark}

\begin{figure}[t]
    \centering
    \includegraphics[width=\linewidth]{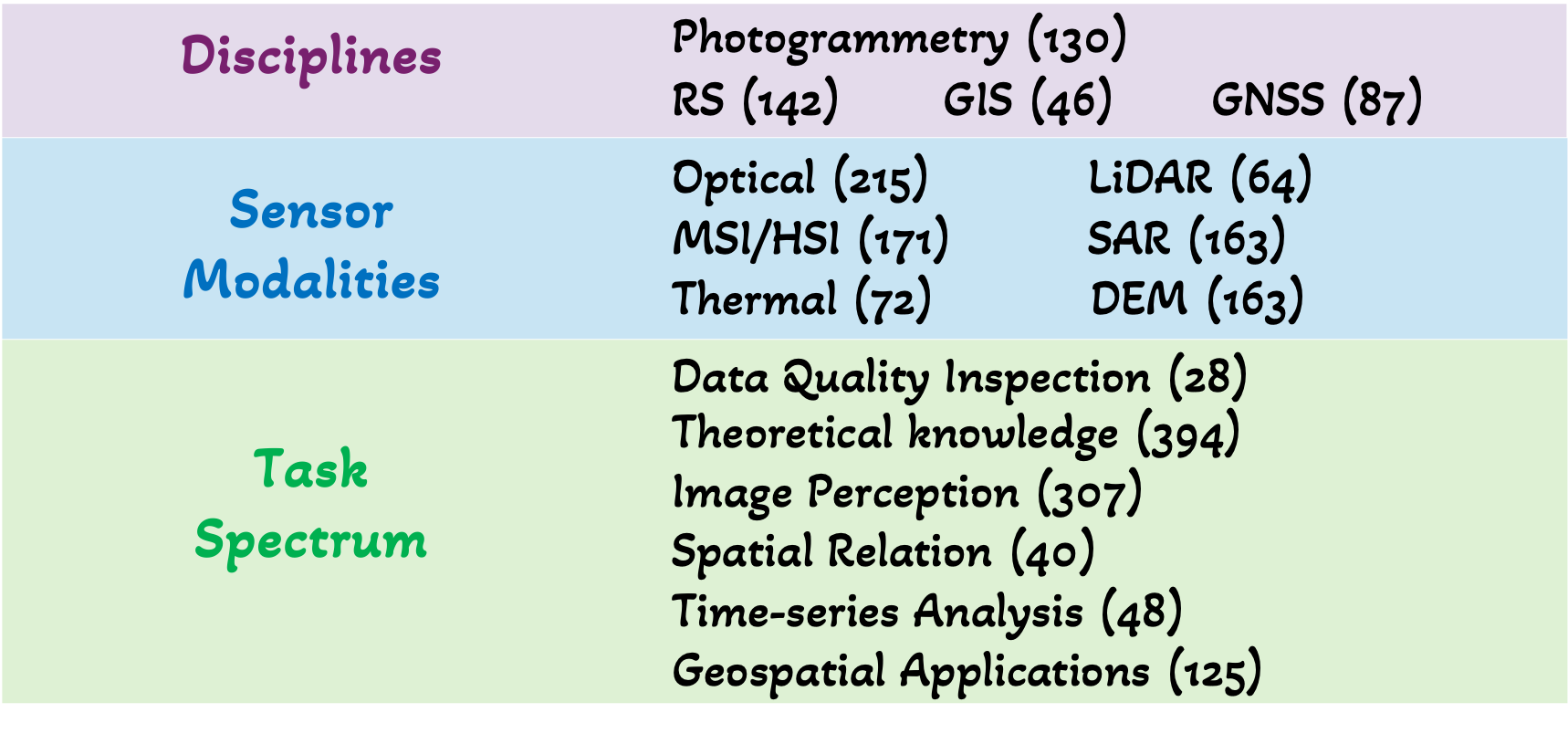}
    \caption{Expert-level knowledge dimensions in geoscience and RS covered in GeoMMBench.}
    \label{fig1:dimensions}
\end{figure}

\begin{figure*}
    \centering
    \includegraphics[width=\linewidth]{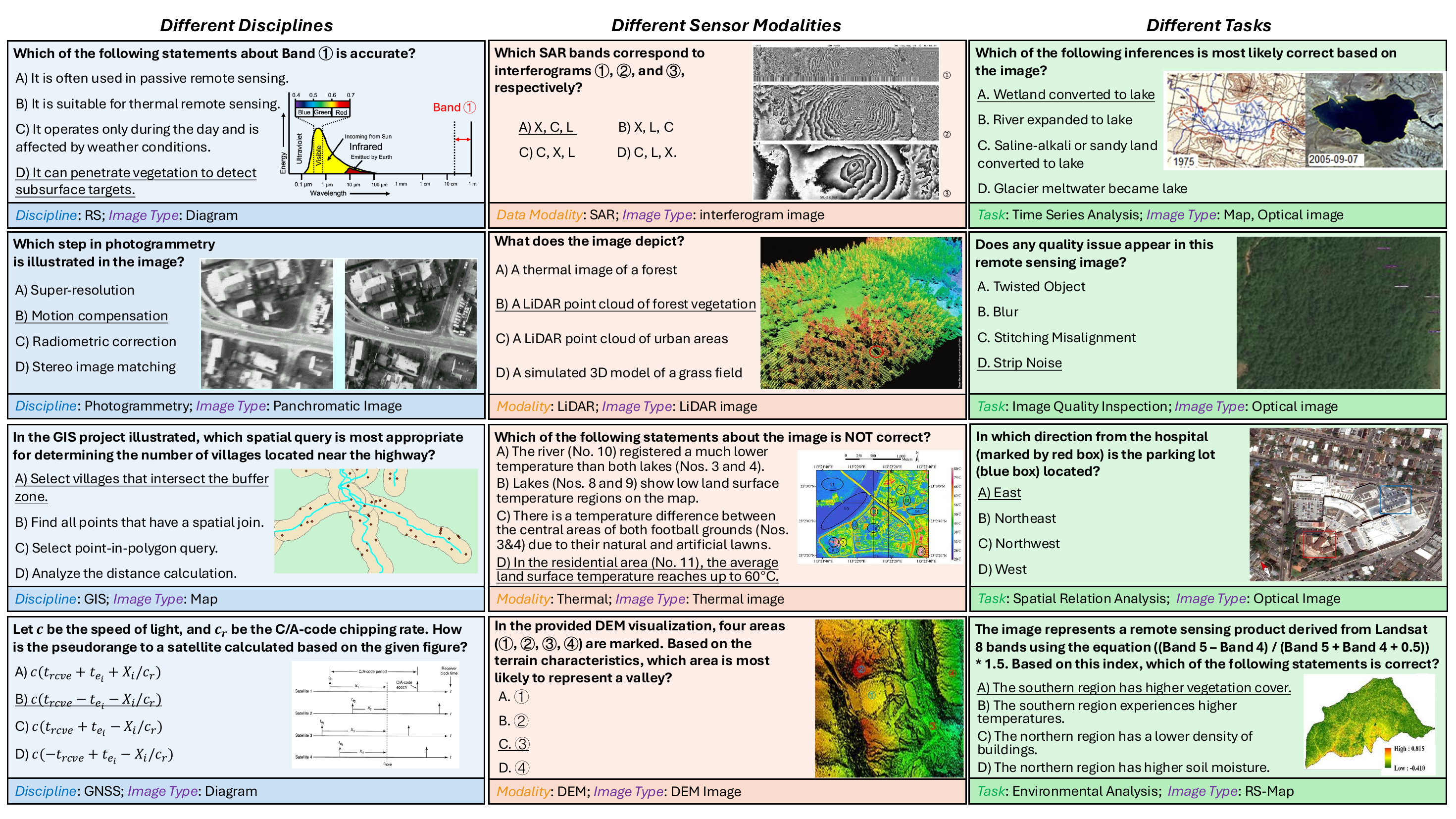}
    \caption{Examples from GeoMMBench, covering multiple disciplines, diverse sensor modalities, and a wide range of task types. Answering these questions requires advanced geoscience and RS expertise, along with specialized reasoning and interpretation skills.}
    \label{fig2:examples}
\end{figure*}

\subsection{Overview}

GeoMMBench is a rigorously designed benchmark that evaluates the ability of MLLMs to comprehend geoscience and RS concepts across multiple dimensions. Its primary objective is to assess how effectively these models perceive and interpret geospatial information across related disciplines while applying domain-specific knowledge and reasoning to solve complex geospatial problems.

The benchmark comprises 1,053 image-based multiple-choice questions, each containing a single correct answer derived from the provided imagery. The dataset was developed through collaborative efforts by domain experts in geoscience and RS, including PhD researchers and doctoral students, who engaged in extensive panel discussions to define the scope, content, and evaluation dimensions. Source materials were drawn from a diverse range of authoritative references, including educational resources, online repositories, and academic literature.

For evaluation purposes, the dataset is divided into a \textit{val} set of 37 questions and a \textit{test} set of 1,016 questions. The \textit{val} set is intended for assessing human expert performance and for model selection or hyperparameter tuning in future studies, whereas the \textit{test} set serves as the final benchmark for performance evaluation. This design ensures a standardized, transparent, and reproducible framework for assessing MLLMs in the geoscience and RS domain.

We highlight key dimensions of GeoMMBench that ensure a specialized and comprehensive evaluation (Fig.~\ref{fig1:dimensions}):

\noindent\textbf{Disciplines:} 
GeoMMBench encompasses four core disciplines: \textit{RS}, \textit{photogrammetry}, \textit{GIS}, and \textit{GNSS}. Advanced geospatial interpretation requires integrating knowledge across these domains. These disciplines are further supported by foundational subjects such as mathematics, physics, geography, cartography, and spectroscopy. The questions frequently integrate multiple disciplinary perspectives to ensure a holistic assessment.

\noindent\textbf{Sensor modalities:}
The benchmark includes a broad range of sensing modalities, such as \textit{MSI/HSI}, \textit{SAR}, \textit{LiDAR}, \textit{DEM}, \textit{optical} and \textit{thermal} imagery. This diversity enables rigorous evaluation of a model’s ability to integrate multi-sensor data for complex Earth observation and geospatial analysis, which is essential for achieving expert-level intelligence.

\begin{table*}[t]
    \centering
    \scriptsize
    \begin{adjustbox}{max width=\textwidth}
    \begin{tabular}{l|cccc|cccccc|cccccc}
    \hline
       \multirow{2}{5em}{Benchmarks} & \multicolumn{4}{c|}{Disciplines} & \multicolumn{6}{c|}{Sensor Modalities} & \multicolumn{6}{c}{Task Spectrum}\\
       & RS & Ph. & GIS & GNSS & Optical & HSI & SAR & DEM &  LiDAR & Thermal & Pri. & Per. & Spa. & Qua. & Tim. & App.\\
    \hline
       VLEO-Bench~\cite{zhang2024good} & \cmark & -- & -- & -- & \cmark & -- & -- & -- & -- & -- & -- & \cmark & -- & -- & \cmark & -- \\
       EarthVQA~\cite{wang2024earthvqa} & \cmark & -- & -- & -- & \cmark & -- & -- & -- & -- & -- & -- & \cmark & -- & -- & \cmark & -- \\
       GeoChat-Bench~\cite{kuckreja2024geochat} & \cmark & -- & -- & -- & \cmark & -- & -- & -- & -- & -- & -- & \cmark & \cmark & -- & -- & -- \\
       VRSBench~\cite{livrsbench} & \cmark & -- & -- & -- & \cmark & -- & -- & -- & -- & -- & -- & \cmark & \cmark & -- & -- & -- \\
       TeoChat-Bench~\cite{irvin2024teochat} & \cmark & -- & -- & -- & \cmark & -- & -- & -- & -- & -- & -- & \cmark & -- & -- & \cmark & -- \\
       LHRS-Bench~\cite{muhtar2024lhrs} & \cmark & -- & -- & -- & \cmark & \cmark & -- & -- & -- & -- & -- & \cmark & \cmark & -- & -- & --\\
       GeoText-1652~\cite{chu2024towards} & \cmark & -- & -- & -- & \cmark & -- & -- & -- & -- & -- & -- & \cmark & -- & -- & -- & -- \\
       GeoBench-VLM~\cite{danish2024geobench}  &  \cmark & -- & -- & -- & \cmark & -- & \cmark & -- & -- & -- & -- & \cmark & \cmark & -- & \cmark & -- \\
       XLRS-Bench~\cite{wang2025xlrs} & \cmark & -- & -- & -- & \cmark & -- & -- & -- & -- & -- & -- & \cmark & \cmark & -- & -- & --\\
        CHOICE~\cite{an2025choice} & \cmark & -- & -- & -- & \cmark & -- & -- & -- & -- & -- & -- & \cmark & \cmark & \cmark & -- & -- \\
    \hline
        GeoMMBench & \cmark & \cmark & \cmark & \cmark & \cmark & \cmark & \cmark & \cmark & \cmark & \cmark & \cmark & \cmark & \cmark & \cmark & \cmark & \cmark \\
    \hline
    \end{tabular}
    \end{adjustbox}
    \caption{Comparison of geoscience and RS benchmarks for MLLMs across key evaluation dimensions. Full names of all dimensions are listed in the Appendix.}
    \label{tab:sum_benchmarks}
\end{table*}

\noindent\textbf{Task spectrum:} GeoMMBench employs a structured task hierarchy to systematically assess key geospatial competencies. It spans four levels: 1) \textit{Theoretical knowledge} across multiple disciplines and subjects, forming the foundation for downstream perception and application tasks. 2) \textit{Low-level perception}, covering image quality assessment and RS-specific corrections (e.g., atmospheric, radiometric, and geometric corrections) for data preprocessing and quality control. 3) \textit{Mid-level geospatial recognition} across sensor modalities, including scene/object recognition, geospatial relationship analysis, and temporal change detection. 4) \textit{High-level geospatial applications} across diverse domains, such as environmental monitoring, agriculture, economic analysis, and RS mapping products.

\noindent\textbf{Image formats:} GeoMMBench incorporates diverse geospatial image formats, including \textit{raw sensor imagery}, \textit{maps}, diagrams, tables, \textit{plots}, and various \textit{RS mapping products}. This diversity challenges models to process heterogeneous visual data while integrating domain-specific knowledge to interpret textual and visual information and apply reasoning skills.

Together, these dimensions ensure that GeoMMBench provides a rigorous and holistic assessment of Expert AGI in geoscience and RS. Fig.~\ref{fig2:examples} presents example questions covering various evaluation dimensions.

\subsection{Benchmark Construction}

\textbf{Data collection.} A team of geoscience and RS experts, including the co-authors, first defined the key evaluation dimensions encompassing disciplines, sensor modalities, and task categories. They then systematically outlined essential knowledge points within these dimensions. Experts were assigned specific knowledge points to develop multimodal questions using major textbooks, online resources, and academic literature, ensuring comprehensive coverage of geospatial expertise by creating new questions when needed. The process strictly adhered to copyright and licensing regulations, ensuring that no data was sourced from restricted materials.

\noindent\textbf{Review process.} 
All questions underwent a rigorous peer review by the expert team. The review process systematically identified and addressed two key issues: 1) Text-only inference bias: Some questions could be answered correctly without relying on the image input, thereby failing to assess true multimodal understanding. These questions were either revised to reinforce image dependency or excluded from the benchmark. 2) Incorrect samples: Questions with ambiguous phrasing, misleading answer choices, or incorrect solutions were identified and refined to uphold benchmark integrity, ensuring that all questions accurately reflected geospatial expertise and aligned with the benchmark’s objectives.

\subsection{Comparisons with Existing Benchmarks}

We elaborate the key differences between GeoMMBench and existing benchmarks from the following perspectives:

\noindent \textbf{Breadth.} Table~\ref{tab:sum_benchmarks} summarizes the evaluation dimensions of GeoMMBench relative to prior benchmarks. Most existing benchmarks are limited in scope, focusing primarily on perception-oriented tasks within the optical domain. GeoMMBench substantially expands this scope by integrating multiple disciplines, diverse sensor modalities, and a comprehensive task spectrum. This broader coverage enables more holistic evaluation of MLLMs and supports progress toward expert-level multimodal intelligence in geoscience.

\noindent\textbf{Depth.} Even when GeoMMBench shares evaluation dimensions with prior benchmarks, its questions demand deeper domain-specific reasoning. As illustrated in Fig.~\ref{fig2:examples}~(row~3, col.~3), GeoMMBench assesses spatial relationship recognition within authentic geoscientific contexts. While earlier benchmarks~\cite{kuckreja2024geochat,danish2024geobench} may include simple spatial queries such as identifying “left” or “right,” GeoMMBench challenges models to determine relative directions using a compass or compute absolute distances from scale bars, which requires advanced spatial reasoning. Similarly, in RS application tasks, models must move beyond basic color recognition in thematic maps to the correct interpretation and use of spectral indices. For instance, GeoMMBench evaluates whether a model can interpret the SAVI\footnote{SAVI: \href{https://www.usgs.gov/landsat-missions/landsat-soil-adjusted-vegetation-index}{Soil-Adjusted Vegetation Index}} and apply it appropriately in a given context (row 4, col. 3). These tasks reflect a substantially deeper level of reasoning that combines conceptual understanding with analytical and application skills, extending well beyond traditional perception-based evaluation.

\begin{figure*}[t]
    \centering
    \includegraphics[width=\linewidth]{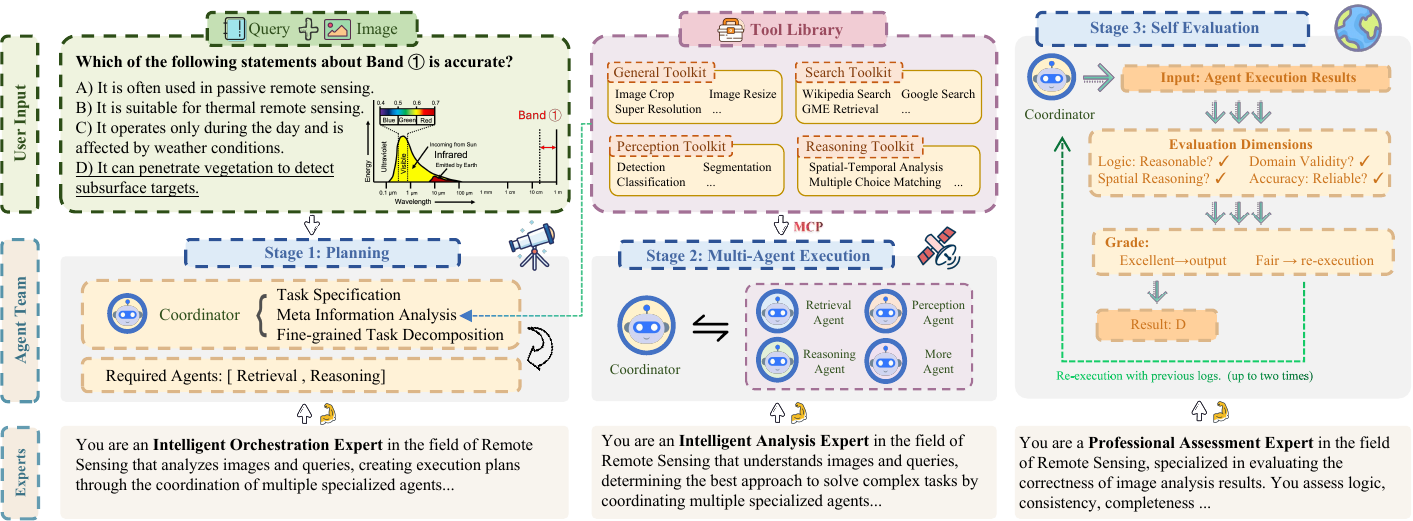}
    \caption{Overview of GeoMMAgent, a multi-agent framework that plans, executes, and self-evaluates multimodal tasks for expert-level geospatial understanding.}
    \label{fig2:method}
\end{figure*}

GeoMMBench remains \textbf{complementary} to existing benchmarks. Prior efforts provide valuable evaluations for optical imagery and foundational vision tasks that support many downstream geospatial applications. Rather than duplicating these contributions, GeoMMBench expands the evaluation landscape toward higher-level reasoning and interpretation skills that are essential for multimodal expert intelligence in geoscience and RS. Additionally, GeoMMBench is \textbf{expert-crafted}, differing fundamentally from prior benchmarks that repurpose perception datasets with text annotations or MLLM-generated questions. Each question is authored by geoscience and RS experts based on domain knowledge and authoritative references, ensuring methodological rigor and substantive depth. This expert-driven construction makes GeoMMBench both uniquely challenging and complementary to existing resources.
\section{GeoMMAgent}

Given the complexity and diversity of geoscience and RS tasks, it is challenging for a single MLLM to meet all domain requirements. A promising solution is to leverage intelligent agent systems that integrate general and specialized tools for adaptive problem-solving. To this end, we introduce GeoMMAgent, a multi-agent orchestration framework designed to enable expert-level interpretation for multimodal tasks in geoscience and RS.

\subsection{Overall Framework}

As illustrated in Fig.~\ref{fig2:method}, GeoMMAgent follows a \textit{plan–execute–evaluate} paradigm comprising three stages: \textit{Planning}, \textit{Multi-Agent Execution}, and \textit{Self-Evaluation}. Given an image–query pair as input, the system dynamically decomposes complex problems into structured subgoals, assigns them to specialized agents, and integrates their outputs into a coherent final answer.

\noindent\textbf{Planning}. A unified coordinator first interprets the input using an internal planner that formulates an executable plan specifying subgoals, required tools or agents, and scheduling strategies for execution order and dependencies.

\noindent \textbf{Multi-Agent Execution.} Based on the execution plan, the coordinator dispatches subtasks to specialized agents responsible for retrieval, perception, and reasoning.
Each agent operates with domain-specific tools, such as object detection or geospatial reasoning models, and returns structured outputs. A comprehensive tool library comprising these specialized agents is described in Section~\ref{sec:tools}. The coordinator then aggregates the results into an interpretable, evidence-based response.

\noindent \textbf{Self-Evaluation.} Finally, a self-evaluation module reviews the reasoning trace and generated answer, verifying consistency and correctness against the original plan. If inconsistencies are detected, the module triggers selective re-execution or refinement.

This modular and iterative workflow enables GeoMMAgent to perform transparent, interpretable, and expert-level analyses for multimodal geoscience and RS tasks.

\subsection{Specialized Tool Library and Core Capabilities}
\label{sec:tools}

We define three core geospatial capabilities that underpin expert-level intelligence: \textit{knowledge}, \textit{perception}, and \textit{reasoning}. These capabilities reflect human expert cognition, as geospatial analysis is inherently complex, dynamic, and multidimensional. Accurate interpretation requires integrating information across multiple sensors, scales, and disciplinary contexts. To operationalize these capabilities, GeoMMAgent provides a tool library that equips agents with domain-specific functionalities for robust interpretation.

\textit{Knowledge tools} handle the retrieval and synthesis of external information. We integrate multiple search engines (e.g., Wikipedia and Google) for general knowledge acquisition and adopt GME~\cite{zhang2025gmeimprovinguniversalmultimodal} for multimodal retrieval to match collected image–text information with the input image–query pair. Retrieved evidence snippets are returned with provenance, ensuring fact-grounded reasoning and expert-level explanations.

\textit{Perception tools} include traditional RS models such as object detection, scene classification, and land use and land cover segmentation, trained with supervised learning on domain datasets. These models are optimized for geospatial perception tasks and provide accurate, interpretable results.

\textit{Reasoning tools} consist of advanced MLLMs that generate contextual insights or intermediate reasoning steps from image–text pairs to support final answer generation. This module is automatically invoked when the coordinator identifies that complex logical structures must be resolved. In our implementation, we employ Qwen-VL-Max, though the framework can accommodate alternative reasoning models.

All tools are implemented in compliance with the Model Context Protocol (MCP)~\cite{hou2025modelcontextprotocolmcp}, ensuring scalability and modularity. This design allows new domain-specific tools to be seamlessly integrated, enabling GeoMMAgent to coordinate heterogeneous agents and perform transparent, expert-level multimodal reasoning in geoscience and RS.
\section{Experiments}
\subsection{Baselines}

\noindent\textbf{Models.} The evaluated models are grouped into three categories. 1) Closed-source MLLMs~\cite{openai2023gpt4vcard,openai2024gpt4o,openai2024o1card,google2024gemini2,team2024gemini} released by commercial providers. 2) Open-source MLLMs further divided into general-purpose~\cite{liu2024improvedbaselinesvisualinstruction,liu2024llavanext,li2024llavaonevisioneasyvisualtask,wang2024cogvlmvisualexpertpretrained,grattafiori2024llama3herdmodels,zhang2024internlm,dong2024internlmxcomposer24khdpioneeringlargevisionlanguage,liu2024nvilaefficientfrontiervisual,chen2024expandingperformanceboundariesopensource,qwen2.5-VL,wang2024qwen2vlenhancingvisionlanguagemodels,abdin2024phi,abouelenin2025phi4minitechnicalreportcompact} and geoscience-specialized variants~\cite{kuckreja2024geochat,irvin2024teochat,muhtar2024lhrs,pang2024vhmversatilehonestvision}. 3) LLMs as text-only baselines~\cite{achiam2023gpt} used to assess the contribution of visual inputs.

\noindent\textbf{Human experts.} 
To establish a human expert baseline, we invited participants with at least a master’s degree in a relevant field to complete the validation tests. They had access to course materials and personal notes but were strictly prohibited from using online search engines or AI assistants.

\begin{table*}[t]
    \centering
    \renewcommand\arraystretch{1.05}
    \begin{adjustbox}{max width=\textwidth}
    \begin{tabular}{l|cc|cccc|cccccc|cccccc}
    \hline
       \multirow{2}{3em}{Model} & \multirow{2}{1em}{Val}  & \multirow{2}{2em}{Test} & \multicolumn{4}{c|}{Results by Disciplines} & \multicolumn{6}{c|}{Results by Sensor Modalities} & \multicolumn{6}{c}{Results by Task Spectrum}\\
       & & & RS & Pho. & GIS & GNS. & Opt. & DEM & SAR & HSI & LiD. & The. & Pri. & Per. & Spa. & Qua. & Tim. & App.\\
    \hline
       \textcolor{gray}{Random}  & \textcolor{gray}{20.8} & \textcolor{gray}{25.2} & \textcolor{gray}{25.3} & \textcolor{gray}{25.2} & \textcolor{gray}{25.0} & \textcolor{gray}{22.4} & \textcolor{gray}{27.2} & \textcolor{gray}{20.4} & \textcolor{gray}{24.5} & \textcolor{gray}{35.1} & \textcolor{gray}{16.1} & \textcolor{gray}{32.1} & \textcolor{gray}{23.2} & \textcolor{gray}{23.7} & \textcolor{gray}{35.1} & \textcolor{gray}{42.3} & \textcolor{gray}{28.3} & \textcolor{gray}{26.3} \\
       Human Expert & 86.5 & \textcolor{gray}{-} & \textcolor{gray}{-} & \textcolor{gray}{-} & \textcolor{gray}{-} & \textcolor{gray}{-} & \textcolor{gray}{-} & \textcolor{gray}{-} & \textcolor{gray}{-} & \textcolor{gray}{-} & \textcolor{gray}{-} & \textcolor{gray}{-} & \textcolor{gray}{-} & \textcolor{gray}{-} & \textcolor{gray}{-} & \textcolor{gray}{-} & \textcolor{gray}{-} & \textcolor{gray}{-} \\
    \hline
        \multicolumn{19}{c}{\textbf{\textit{LLMs: Only Text as Input}}} \\
    \hline
        GPT-4 Text \cite{achiam2023gpt} & 24.3 & 31.2 & 27.4 & 49.6 & 36.4 & 45.9 & 19.9 & 29.9 & 26.4 & 32.4 & 35.5 & 35.7 & 30.8 & 27.1 & 16.2 & 7.7 & 30.4 & 30.5\\
    \hline
        \multicolumn{19}{c}{\textbf{\textit{Open-source MLLMs}}} \\
    \hline
        GeoChat~\cite{kuckreja2024geochat} & 21.6 & 35.4 & 34.1 & 40.9 & 34.1 & 34.1 & 41.7 & 22.8 & 34.6 & 27.0 & 46.8 & 21.4 & 31.1 & 36.8 & 29.7 & 26.9 & 30.4 & 33.9\\
        TeoChat~\cite{irvin2024teochat} & 24.3 & 31.1 & 28.2 & 39.4 & 52.3 & 37.6 & 29.6 & 22.8 & 30.8 & 32.4 & 25.8 & 25.0 & 29.2 & 26.4 & 27.0 & 23.1 & 32.6 & 28.0\\
        SkySenseGPT \cite{luo2024sky} & 18.9 & 36.5 & 35.8 & 40.2 & 36.4 & 37.6 & 48.1 & 25.7 & 31.4 & 37.8 & 46.8 & 25.0 & 30.3 & 42.5 & 37.8 & 26.9 & 28.3 & 30.5\\
        LHRS-Bot \cite{muhtar2024lhrs} & 27.0 & 31.1 & 27.9 & 37.0 & 45.5 & 41.2 & 31.6 & 26.9 & 25.2 & 29.7 & 21.0 & 25.0 & 24.0 & 33.8 & 13.5 & 15.4 & 32.6 & 33.1\\
        VHM~\cite{pang2024vhmversatilehonestvision} & 37.8 & 39.1 & 39.0 & 40.2 & 27.3 & 43.5 & 56.3 & 28.1 & 30.8 & 35.1 & 41.9 & 35.7 & 27.7 & 46.2 & 35.1 & 38.5 & 41.3 & 44.9 \\
    \hdashline
        LLaVA1.5-7B~\cite{liu2024improvedbaselinesvisualinstruction} & 27.0 & 34.4 & 34.6 &  7.6 & 45.5 & 27.1 & 46.6 & 21.0 & 38.4 & 27.0 & 41.9 & 32.1 & 29.2 & 40.8 & 37.8 & 34.6 & 26.1 & 30.5 \\
        LLaVA1.5-13B~\cite{liu2024improvedbaselinesvisualinstruction} & 29.7 & 38.2 & 38.9 & 32.3 & 50.0 & 27.1 & 57.3 & 27.5 & 37.7 & 32.4 & 32.3 & 28.6 & 32.4 & 45.5 & 45.9 & 34.6 & 41.3 & 35.6 \\
        LLaVA-NeXT-7B~\cite{liu2024llavanext} & 45.9 & 41.2 & 42.0 & 40.2 & 54.5 & 35.3 & 54.9 & 31.7 & 37.1 & 32.4 & 40.3 & 53.6 & 36.8 & 47.8 & 51.4 & 34.6 & 39.1 & 39.8 \\
        LLaVA-NeXT-34B~\cite{liu2024llavanext} & 59.5 & 54.2 & 52.8 & 63.0 & 56.8 & 60.0 & 60.2 & 42.5 & 52.2 & 51.4 & 74.2 & 35.7 & 50.7 & 51.8 & 59.5 & 46.2 & 54.3 & 55.9\\
        LLaVA-OneVision-0.5B~\cite{li2024llavaonevisioneasyvisualtask} & 32.4 & 40.3 & 39.9 & 44.1 & 40.9 & 35.3 & 48.5 & 32.3 & 35.8 & 35.1 & 51.6 & 32.1 & 35.0 & 41.8 & 35.1 & 46.2 & 37.0 & 41.5\\
        LLaVA-OneVision-7B~\cite{li2024llavaonevisioneasyvisualtask} & 56.8 & 57.3 & 55.7 & 64.6 & 65.9 & 64.7 & 66.5 & 50.3 & 47.8 & 59.5 & 62.9 & 46.4 & 53.5 & 57.5 & 54.1 & \textbf{65.4} & 52.2 & 51.7\\
        CogVLM2~\cite{wang2024cogvlmvisualexpertpretrained} & 37.8 & 54.1 & 52.8 & 55.9 & 54.5 & 60.0 & 67.0 & 48.5 & 43.4 & 62.2 & 61.3 & 53.6 & 47.0 & 57.9 & 40.5 & 57.7 & 54.3 & 54.2 \\
        Mini-Monkey~\cite{huang2024mini} & 54.1 & 41.8 & 41.4 & 37.0 & 52.3 & 43.5 & 48.1 & 34.7 & 36.5 & 43.2 & 50.0 & 32.1 & 37.1 & 43.8 & 37.8 & 42.3 & 45.7 & 43.2\\
        LLaMA3.2-11B~\cite{grattafiori2024llama3herdmodels} & 35.1 & 40.1 & 38.1 & 46.5 & 50.0 & 43.5 & 46.1 & 32.9 & 35.8 & 43.2 & 45.2 & 28.6 & 37.1 & 40.1 & 29.7 & 42.3 & 30.4 & 36.4\\
        InternLM-XComposer-2.5-7B~\cite{zhang2024internlm} & 37.8 & 46.2 & 45.3 & 47.2 & 63.6 & 45.9 & 54.4 & 37.7 & 37.7 & 70.3 & 48.4 & 50.0 & 39.9 & 48.5 & 35.1 & 46.2 & 54.3 & 48.3 \\
        InternLM-XComposer2-4KHD~\cite{dong2024internlmxcomposer24khdpioneeringlargevisionlanguage} & 44.5 & 31.3 & 31.6 & 29.1 & 43.2 & 17.6 & 52.9 & 25.1 & 22.6 & 27.0 & 29.0 & 28.6 & 24.3 & 40.5 & 59.5 & 34.6 & 23.9 & 22.0 \\
        NVILA-8B~\cite{liu2024nvilaefficientfrontiervisual} & 43.2 & 45.3 & 42.4 & 57.5 & 47.7 & 55.3 & 34.5 & 45.5 & 37.7 & 67.6 & 56.5 & 46.4 & 42.8 & 39.5 & 35.1 & 30.8 & 45.7 & 51.7\\
        NVILA-15B~\cite{liu2024nvilaefficientfrontiervisual} & 48.6 & 46.8 & 43.7 & 62.2 & 47.7 & 56.5 & 30.6 & 47.3 & 49.7 & 56.8 & 59.7 & 50.0 & 48.3 & 35.8 & 37.8 & 42.3 & 45.7 & 57.6\\
        InternVL 2.5-8B~\cite{chen2024expandingperformanceboundariesopensource} & 59.5 & 56.8 & 55.2 & 61.4 & 65.9 & 62.4 & 65.0 & 46.1 & 42.1 & 59.5 & 64.5 & 53.6 & 50.4 & 55.2 & 54.1 & 57.7 & 60.9 & 64.4 \\
        Phi-3v-128k \cite{abdin2024phi} & 45.9 & 46.9 & 45.2 & 51.2 & 63.6 & 51.8 & 51.5 & 37.1 & 44.0 & 51.4 & 54.8 & 39.3 & 44.1 & 44.8 & 37.8 & 38.5 & 39.1 & 47.5\\
        Phi-3.5v \cite{abdin2024phi} & 51.4 & 48.7 & 46.8 & 55.1 & 63.6 & 60.0 & 50.5 & 39.5 & 40.3 & 54.1 & 54.8 & 46.4 & 46.2 & 44.5 & 54.1 & 34.6 & 47.8 & 47.5\\
        Phi-4-Multimodal \cite{abouelenin2025phi4minitechnicalreportcompact} & 48.6 & 55.3 & 54.0 & 63.8 & 61.4 & 62.4 & 58.3 & 50.3 & 47.2 & 70.3 & 61.3 & 50.0 & 53.3 & 53.2 & 40.5 & \textbf{65.4} & 47.8 & 58.5\\
        Qwen2-VL-2B~\cite{wang2024qwen2vlenhancingvisionlanguagemodels} & 29.7 & 48.9 & 48.2 & 48.8 & 59.1 & 49.4 & 61.7 & 39.5 & 39.0 & 54.1 & 53.2 & 42.9 & 43.9 & 49.2 & 51.4 & 61.5 & 47.8 & 47.5 \\
        Qwen2-VL-7B~\cite{wang2024qwen2vlenhancingvisionlanguagemodels} & 51.4 & 59.8 & 59.1 & 56.7 & 70.5 & 61.2 & 69.4 & 51.5 & 54.1 & 67.6 & 67.7 & 50.0 & 53.8 & \textbf{63.5} & \textbf{62.2} & 46.2 & 58.7 & 61.9 \\
        Qwen2.5-VL-3B~\cite{qwen2.5-VL} & 54.1 & 58.1 & 56.8 & 60.6 & 61.4 & 63.5 & 64.1 & 48.5 & 52.8 & 62.2 & 62.9 & 60.7 & 54.3 & 57.9 & 59.5 & 42.3 & 52.2 & 57.6\\
        Qwen2.5-VL-7B~\cite{qwen2.5-VL} & 56.8 & 62.5 & 60.6 & 68.5 & 72.7 & 70.6 & 66.0 & 52.1 & 55.3 & 70.3 & 74.2 & 57.1 & 59.8 & 59.2 & 54.1 & 50.0 & 52.2 & 64.4\\
        Qwen3-VL-4B \cite{qwen3technicalreport} & 56.8 & 63.2 & 61.2 & 72.4 & 75.0 & 68.2 & \textbf{73.8} & 49.7 & 52.2 & 73.0 & 75.8 & 60.7 & 58.2 & 58.9 & 56.8 & 53.8 & 58.7 & 64.4\\
        Qwen3-VL-8B \cite{qwen3technicalreport} & 56.8 & 65.8 & 64.0 & 70.1 & \textbf{77.3} & 70.6 & 72.8 & 59.3 & 55.3 & \textbf{75.7} & 75.8 & \textbf{67.9} & 61.4 & 62.5 & 51.4 & 57.7 & 60.9 & \textbf{71.2} \\
        Qwen3-VL-30B \cite{qwen3technicalreport} & \textbf{70.3} & \textbf{66.7} & \textbf{64.5} & \textbf{74.0} & 72.7 & \textbf{76.5} & 71.4 & \textbf{59.9} & \textbf{58.5} & 70.3 & \textbf{80.6} & 60.7 & \textbf{63.2} & 61.2 & 48.6 & \textbf{65.4} & \textbf{63.0} & 67.8 \\
    \hline
        \multicolumn{19}{c}{\textbf{\textit{Closed-source MLLMs}}} \\
    \hline
        GPT-4v \cite{openai2023gpt4vcard} & 56.8 & 61.0 & 59.3 & 71.7 & 61.4 & 67.1 & 63.6 & 61.7 & 54.1 & 45.9 & 71.0 & 46.4 & 59.8 & 58.5 & 37.8 & 53.8 & 56.5 & 69.5\\
        GPT-4o \cite{openai2024gpt4o} & 56.8 & 68.5 & 65.6 & 78.7 & 77.3 & \textbf{80.0} & 71.4 & 62.3 & 61.0 & \textbf{64.9} & \textbf{80.6} & 67.9 & 65.8 & 63.9 & 48.6 & 69.2 & 56.5 & \textbf{72.0}\\
        GPT-o1 \cite{openai2024o1card} & 59.5 & 65.8 & 63.5 & 78.7 & 65.9 & 72.9 & 71.4 & 63.5 & 56.0 & 59.5 & 75.8 & 67.9 & 64.2 & 63.5 & 54.1 & 50.0 & 56.5 & 64.4 \\
        GPT-5 \cite{openai_gpt5_systemcard_2025} & 35.1 & 46.1 & 41.2 & 64.6 & 54.5 & 67.1 & 21.8 & 46.1 & 39.6 & 48.6 & 61.3 & 28.6 & 46.0 & 33.8 & 21.6 & 30.8 & 50.0 & 59.3\\
        Gemini-1.5-pro \cite{team2024gemini} & \textbf{64.8} & \textbf{70.7} & \textbf{68.9} & 76.4 & \textbf{86.4} & 72.9 & \textbf{74.3} & 64.7 & \textbf{67.3} & \textbf{64.9} & 77.4 & \textbf{75.0} & \textbf{71.5} & \textbf{68.2} & 56.8 & 61.5 & 58.7 & 62.7 \\
        Gemini-2.0-flash \cite{google2024gemini2} & 62.2 & 70.1 & 67.5 & \textbf{79.5} & \textbf{86.4} & 74.1 & 73.8 & \textbf{67.1} & 62.3 & \textbf{64.9} & 72.6 & \textbf{75.0} & 68.9 & 66.2 & \textbf{59.5} & \textbf{69.2} & \textbf{63.0} & 64.4 \\
    \hline
        \textbf{GeoMMAgent (ours)} & \textbf{86.5} & \textbf{88.4} & \textbf{87.6} & \textbf{89.8} & \textbf{93.2} & \textbf{97.6} & \textbf{78.2} & \textbf{94.0} & \textbf{91.2} & \textbf{89.2} & \textbf{74.2} & \textbf{89.3}& \textbf{92.2} & \textbf{82.9} & \textbf{97.3} & \textbf{76.9} & \textbf{97.8} & \textbf{83.1}\\
    \hline
    \end{tabular}
    \end{adjustbox}
    \vspace{-5pt}
    \caption{Evaluation results of MLLMs and text-only LLM baselines on the GeoMMBench \textit{val} and \textit{test} sets. The best-performing model in each category is highlighted in bold. Abbreviations: Full names of all dimensions and their subtasks are listed in the Appendix.}
    \label{tab:sota_performance}
\end{table*}

\begin{figure*}[t]
    \centering
    \includegraphics[width=\linewidth]{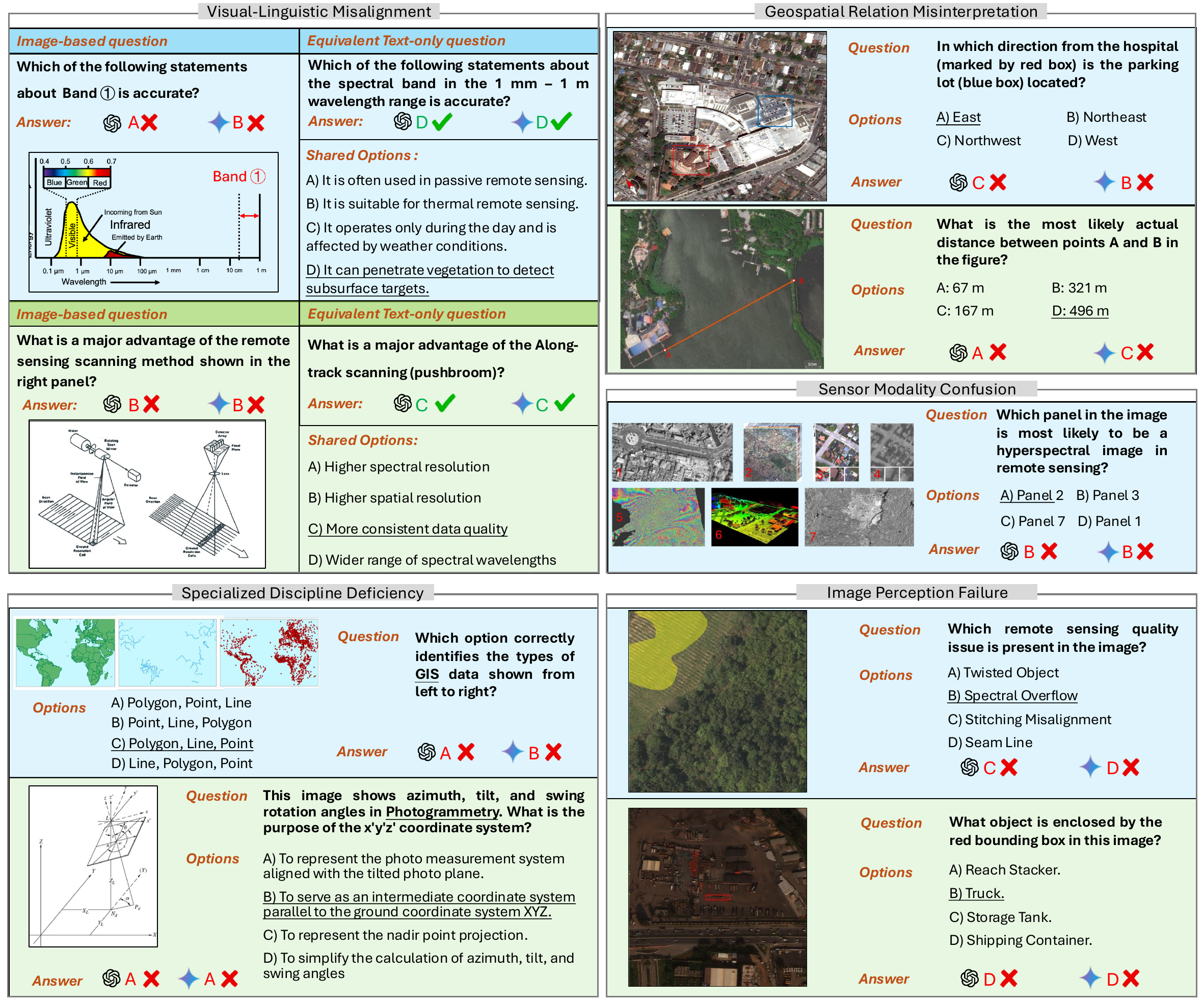}
    \vspace{-15pt}
    \caption{Representative error cases from advanced MLLMs specific to geospatial tasks. \includegraphics[height=8pt]{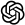}: GPT-4o; \includegraphics[height=8pt]{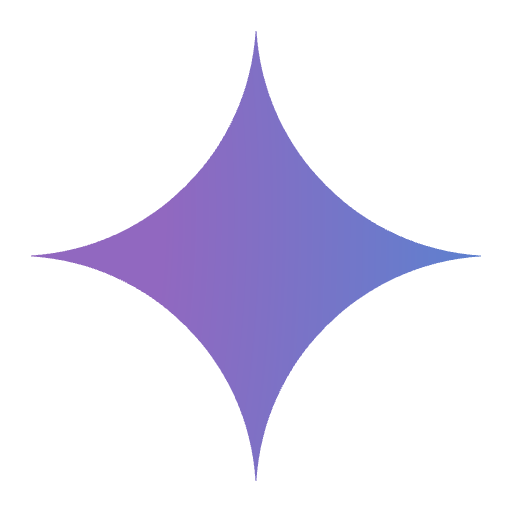}: Gemini-1.5 pro.}
    \label{fig:false_cases}
\end{figure*}

\noindent \textbf{Evaluation.} All models, including GeoMMAgent, are tested under a zero-shot setting without fine-tuning or additional supervision. Each model follows its default multiple-choice question-answering prompt for consistency. Model responses are evaluated through an automated rule-based pipeline using regular expressions to extract option letters and numerical answers, with invalid outputs marked as incorrect. Micro-averaged accuracy serves as the primary metric. All evaluations are conducted using the LMMs-Eval toolkit~\cite{zhang2024lmmsevalrealitycheckevaluation}, with random-choice results reported as a baseline. Experiments are performed on NVIDIA A100 GPUs.

\subsection{Main Results}

We present the evaluation results in Table~\ref{tab:sota_performance} and summarize our key findings as follows:

\noindent\textbf{GeoMMBench poses significant challenges for current models.} 
Human experts achieve a validation accuracy of 86.5\%, outperforming all foundation models by a large margin. This pronounced gap underscores the limitations of existing models in geoscience and RS, revealing weaknesses in expert-level reasoning, multimodal integration, and domain-specific interpretation. These results demonstrate the rigor of GeoMMBench and its effectiveness in evaluating advanced geospatial competencies.

\noindent\textbf{GeoMMAgent demonstrates superior capability.}
GeoMMAgent attains 88.4\% accuracy on the test set, surpassing all MLLMs by substantial margins. It also achieves the same score by human expert on val set. 
This is due to its multi-agent design, enabling the intelligent decomposition of complex RS tasks and coordinated use of specialized tools, highlighting the great potential of agent-based systems for expert-level multimodal analysis in RS.

\noindent\textbf{Open-source MLLMs are rapidly improving.}
State-of-the-art closed-source models still lead overall, with Gemini-1.5 Pro \cite{team2024gemini} reaching 70.7\% accuracy. The leading open-source model, Qwen3-VL-30B \cite{qwen3technicalreport}, achieves 66.7\%, narrowing the gap and even outperforming GPT-4o \cite{openai2024gpt4o}. Given the importance of geoscience and RS for public and sensitive applications, continued progress in open-source MLLMs is critical for accessibility and private deployment.

\noindent\textbf{Geospatial-specific models require stronger instruction tuning.}
Domain-specialized MLLMs such as GeoChat \cite{kuckreja2024geochat}, TeoChat \cite{irvin2024teochat}, LHRS-Bot \cite{muhtar2024lhrs}, and VHM \cite{pang2024vhmversatilehonestvision} perform markedly below advanced general-purpose models despite fine-tuning on RS data. Their narrow focus on perception tasks limits reasoning and generalization in broader geospatial contexts. These results highlight the need for more diverse datasets and instruction-tuning strategies to enhance domain-specific expertise.

\subsubsection{Error Analysis}
To better understand the limitations of SOTA MLLMs in geoscience and RS, we analyze misinterpreted samples from Gemini-1.5 Pro and GPT-4o, the top-performing models from Google and OpenAI, respectively, as reported in Table~\ref{tab:sota_performance}. Fig.~\ref{fig:false_cases} presents representative error cases specific to geospatial tasks, highlighting common geo-interpretation failure patterns in model predictions. These cases underscore the need for domain-specialized advancements in future developments.

\noindent\textbf{Visual-linguistic misalignment.} A key challenge in evaluating Expert AGI for geoscience and RS is distinguishing between genuine gaps in domain knowledge and failures caused by misalignment between language and vision components. Our analysis reveals that while SOTA LLMs often exhibit strong geospatial knowledge in text-based tasks, their visual understanding remains inadequate, resulting in incorrect predictions.
Fig.~\ref{fig:false_cases} (top-left) illustrates two representative cases. When given text-only input, GPT-4o and Gemini-1.5 Pro accurately describe the spectral characteristics of 1mm–1nm wavelengths (i.e., microwave) and recognize their ability to penetrate vegetation for subsurface detection. However, when required to identify the wavelength range for Band \textcircled{1} in a standard electromagnetic spectrum diagram, both models fail, selecting incorrect options despite possessing the necessary textual knowledge. A similar issue arises with Along-Track Scanning (Pushbroom) sensors. Both models correctly explain their advantages in a text-based setting but fail to recognize them visually, leading to misinterpretation.

These findings highlight a key limitation: while MLLMs may exhibit rich geospatial knowledge in text-based tasks, their visual encoders remain poorly aligned, preventing effective integration of visual and textual reasoning. Bridging this gap requires improving visual processing architectures to enhance the interpretation of domain-specific imagery in geoscience and RS.

\noindent\textbf{Misinterpretation of geospatial relationships.} Interpreting spatial relationships is fundamental in geoscience and RS, often requiring expertise in analyzing spatial elements such as compasses (for direction), scale bars (for distance), and latitude/longitude grids (for positioning). GeoMMBench evaluates these geospatial reasoning skills through a diverse set of targeted questions. As shown in Fig.~\ref{fig:false_cases} (upper right), existing MLLMs struggle with complex spatial reasoning, often failing to interpret relationships correctly, even when given clear visual cues such as legends. These findings highlight a critical gap in domain-specific spatial reasoning and emphasize the need for geospatial-aware MLLMs that can better integrate visual, numerical, and contextual information for precise spatial interpretation.

\noindent\textbf{Sensor modality confusion.} As shown in Fig.~\ref{fig:false_cases} (middle right), MLLMs may misinterpret images from different sensor types, failing to distinguish between various RS modalities. This challenge stems from training data bias, as most models are predominantly trained on RGB imagery, limiting their exposure to non-RGB RS such as HSI imagery. Overcoming this limitation requires expanding training datasets to include a more diverse and representative range of RS sensor modalities. This improvement will enhance models’ multimodal understanding, strengthening their ability to perform complex geospatial analyses and fostering synergistic multi-sensor processing.

\noindent\textbf{Deficiencies in specialized geoscience disciplines.} MLLMs often struggle with less common geoscience disciplines, likely due to their limited representation in training data. As shown in Fig.~\ref{fig:false_cases} (bottom left), the sample questions cover fundamental concepts in GIS and photogrammetry—topics considered standard knowledge in the field. However, even the two best-performing proprietary MLLMs fail to answer them correctly, revealing significant gaps in their domain understanding. This highlights a critical gap in existing models and underscores the need for specialized geospatial MLLMs with deeper domain knowledge to improve reasoning and interpretation in geoscience and RS.

\noindent\textbf{Image perception failure.} While GeoMMBench does not primarily focus on perception tasks for raw RS images like prior benchmarks (as discussed in Section 3.3), it incorporates domain-specific recognition tasks that are essential for geoscience applications. Our evaluation indicates that while advanced MLLMs perform well in coarse-level recognition tasks for RGB images (e.g., scene classification), they struggle with more sophistic recognition tasks, as shown in Fig.~\ref{fig:false_cases} (bottom right). These challenges span from low-level assessments, such as RS-specific image quality inspection and correction, to mid-level tasks, including fine-grained object recognition. Since these recognition capabilities serve as the foundation for intelligent RS interpretation, improving MLLMs' ability to handle domain-specific geospatial perception tasks remains a crucial direction for future research and development in geospatial AI.
\section{Conclusion}
\vspace{-5pt}
In this work, we introduced GeoMMBench, the first expert-crafted, multidisciplinary benchmark for evaluating MLLMs in geoscience and RS. Through comprehensive experiments, we revealed substantial challenges that current models face in expert-level reasoning, perception, and multimodal integration. To address these limitations, we developed GeoMMAgent, a multi-agent orchestration framework that integrates domain-specific tools for adaptive task planning and expert-level interpretation. GeoMMAgent achieves state-of-the-art performance on GeoMMBench, surpassing all existing models and approaching human expert performance. We believe that GeoMMBench and GeoMMAgent together lay a foundation for advancing intelligent, transparent, and domain-specialized multimodal agents in geoscience and beyond.

\section*{Acknowledgement}
This work was supported in part by JST FOREST (Grant Number JPMJFR206S) and JST CRONOS (Grant Number JPMJCS25K5). It was also supported in part by the Excellence Center at Linköping-Lund in Information Technology (ELLIIT) Researcher Funding, the Zenith Research Program, and the Swedish Research Council with grant agreement No. 2024-05652. Additionally, it was supported in part by the National Natural Science Foundation of China (NSFC) under Grant No. 62401273.

{
    \small
    \bibliographystyle{ieeenat_fullname}
    \bibliography{main}
}

\setcounter{page}{1}
\maketitlesupplementary

\renewcommand{\thefigure}{A\arabic{figure}}
\renewcommand{\thetable}{A\arabic{table}}
\setcounter{figure}{0}
\setcounter{table}{0}

\section{More Descriptions on GeoMMBench}
\subsection{Dimensions and Tasks in GeoMMBench}
Below we provide explanations for the abbreviations of evaluation dimensions in GeoMMBench, as listed in Tables 1 and 3 of the paper, along with their corresponding tasks.

\noindent \textbf{Disciplines:} “RS” (Remote Sensing), “Ph.” (Photogrammetry), “GIS” (Geographic Information System), and “GNSS” (Global Navigation Satellite System). Please note that in this paper, RS in “geoscience and remote sensing” is used in a broad sense, referring to the observation of objects without physical contact. At the disciplinary level, photogrammetry focuses on geometric measurements, whereas RS is primarily concerned with radiation-based measurements.

\noindent \textbf{Sensor Modalities:} “Opt.”: Optical RGB imagery, “HSI”: Multispectral/Hyperspectral imagery (MSI can be regarded as a special case of HSI with fewer spectral bands, so we simplify here for briefly), “SAR”: Synthetic Aperture Radar, “LiD.”: Light Detection and Ranging, “DEM”: Digital Elevation Model, “The.”: Thermal imagery.

\noindent \textbf{Task Spectrum:} 
\begin{itemize}
    \item “Pri.” (Principles) refers to fundamental principles analysis and theoretical understanding.
	\item “Per.” (Perception) covers perception tasks across different sensor modalities, including: \textit{Basic perception tasks} (scene classification, object referring, object grounding, object counting, and sensor-level recognition such as modality discrimination); \textit{Fine-level object recognition} and \textit{complex interpretation of sensor imagery}, including attribute analysis and identification/interpretation of spectral curves.
	\item “Spa.” (Spatial) refers to spatial relation analysis, including relative direction, distance estimation, spatial positioning, and other geospatial interpretations such as estimating ocean depth or mountain height.
	\item “Qua.” (Quality) includes data quality tasks such as image quality inspection and corrections (atmospheric, radiometric, and geometric).
	\item “Tim.” (Time Series) refers to temporal analysis across multiple images with various formats, including sensor imagery, maps, and RS mapping products.
	\item “App.” (Applications) covers RS applications, focusing on the analysis of RS products across various domains, such as economics, environment, oceanography, and climate studies.
\end{itemize}

\subsection{Different Image Types.}

\begin{figure}[h]
    \centering
    \includegraphics[width=\linewidth]{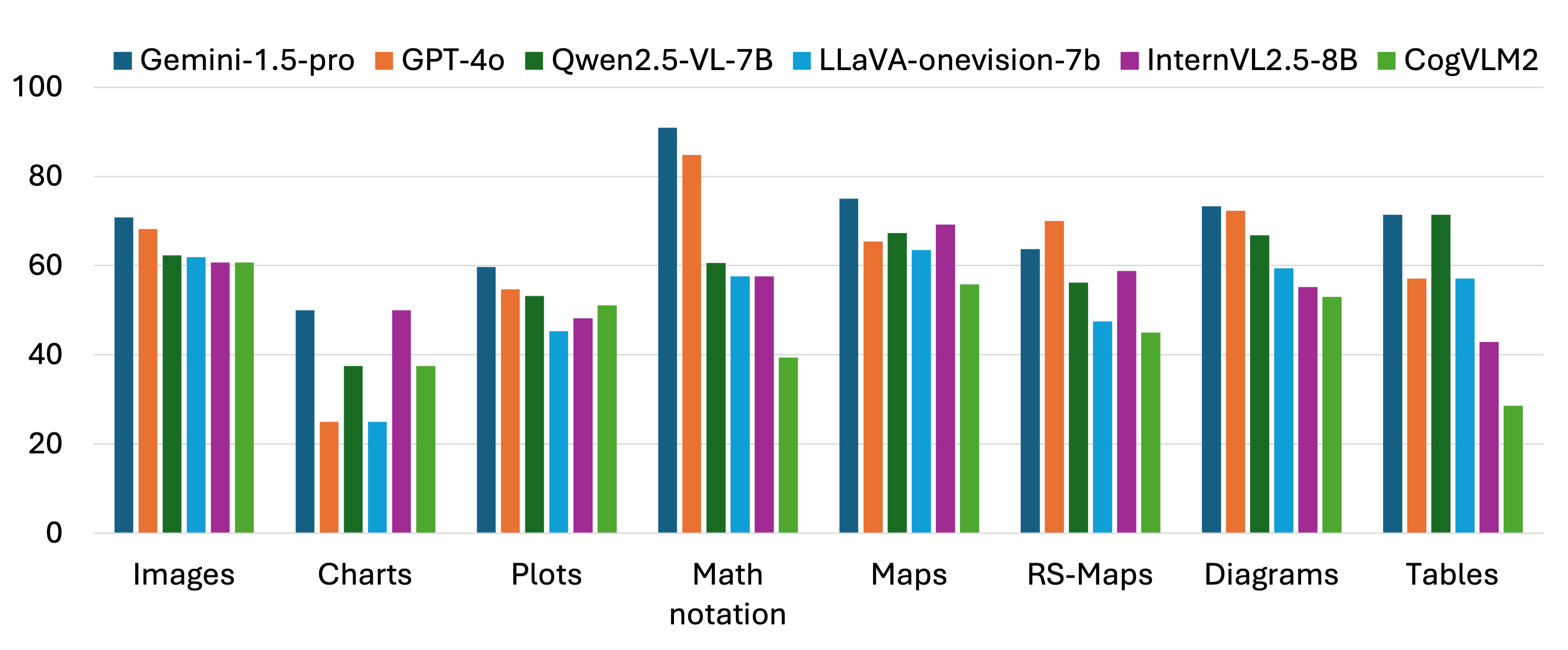}
    \caption{Performance of MLLMs on different types of images in GeoMMExpert.}
    \label{fig:image_types}
\end{figure}

We analyze model performance across various image types in Fig. \ref{fig:image_types}, including sensor imagery, charts, plots, mathematical notations, maps, RS-map products, diagrams, and tables. Among all models, Gemini-1.5 Pro demonstrates the highest accuracy across all categories, suggesting its robust generalization to diverse visual formats commonly used in geoscience and RS. Open-source models, while improving in some areas, exhibit notable weaknesses in mathematical notation interpretation and show relatively lower performance in charts and RS-map products. This indicates that they face difficulties with these specialized representations, which require both multimodal understanding and domain-specific reasoning. These results highlight the importance of enhancing model adaptability to structured and geospatial-specific visual information for improved expert-level AGI performance.

\subsection{More Error Cases}

In this subsection, we provide additional error cases from GPT-4o~\cite{openai2024gpt4o}, as shown in Figs.~\ref{fig:1} to \ref{fig:15}. These cases span diverse geoscience and RS dimensions, illustrating common failure patterns in SOTA MLLMs when interpreting geospatial data and tasks.

\begin{figure}[h]
    \centering
    \includegraphics[width=\linewidth]{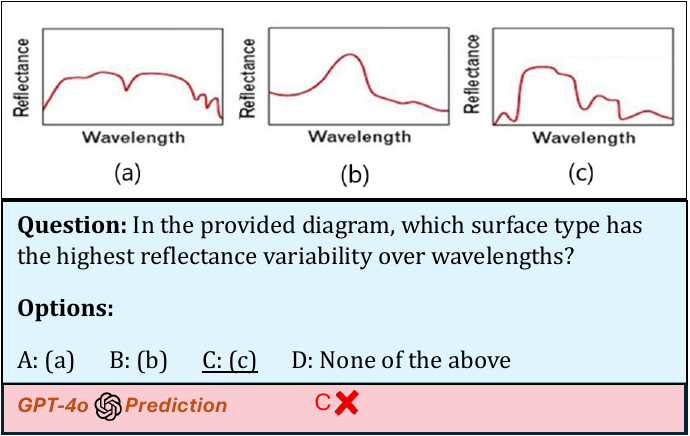}
    \vspace{-20pt}
    \caption{A sample error case of remote sensing principle understanding. Subfield: Spectral curve interpretation.}
    \label{fig:enter-label}
    \vspace{-10pt}
\end{figure}

\begin{figure}[h]
    \centering
    \includegraphics[width=\linewidth]{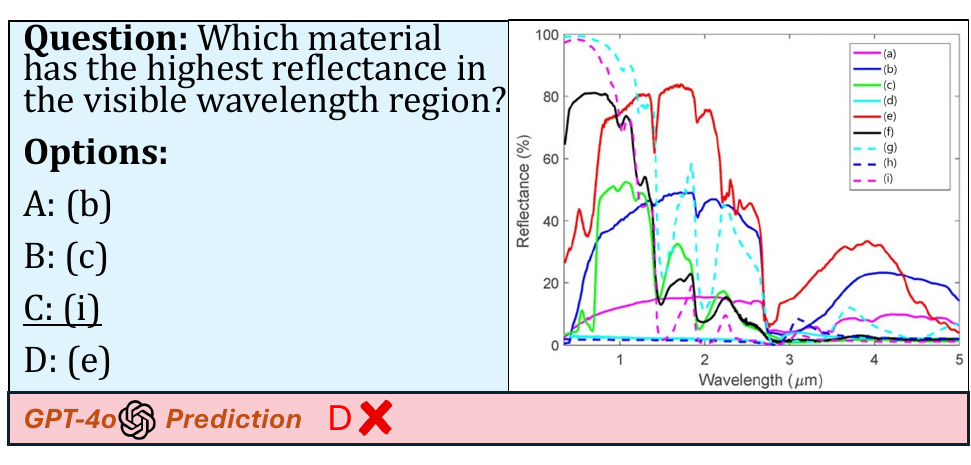}
    \vspace{-5pt}
    \caption{A sample error case of remote sensing principle understanding. Subfield: Spectral curve interpretation.}
    \vspace{-5pt}
\end{figure}

\begin{figure}[h]
    \centering
    \includegraphics[width=\linewidth]{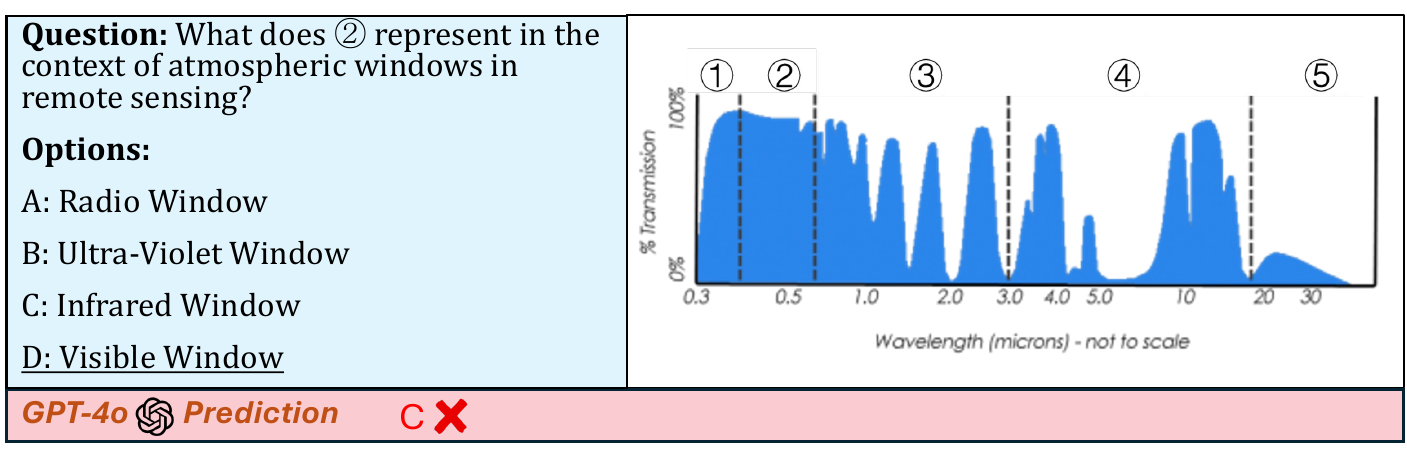}
    \caption{A sample error case in remote sensing principles. Subfield: Atmospheric window recognition.}
    \label{fig:1}
\end{figure}

\begin{figure}[h]
    \centering
    \includegraphics[width=\linewidth]{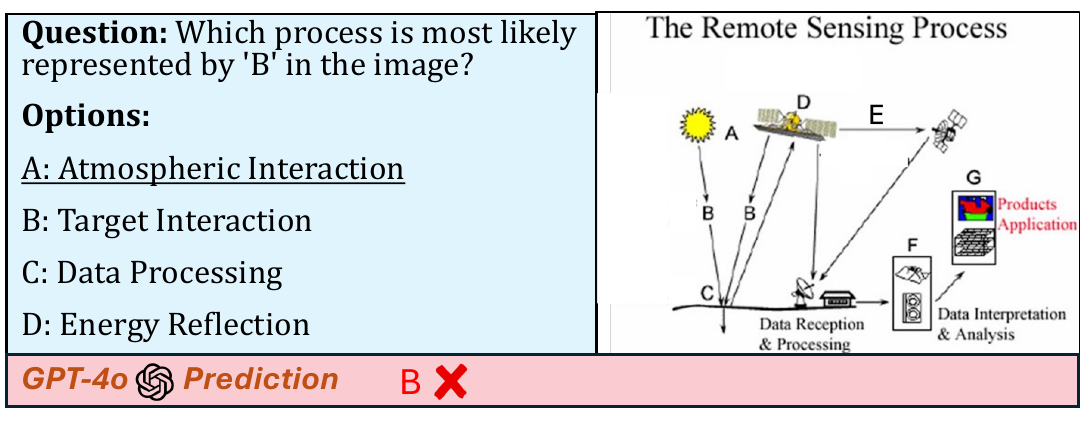}
    \vspace{-20pt}
    \caption{A sample error case in remote sensing principles. Subfield: Remote sensing process.}
    \label{fig:enter-label}
    \vspace{-10pt}
\end{figure}

\begin{figure}[h]
    \centering
    \includegraphics[width=\linewidth]{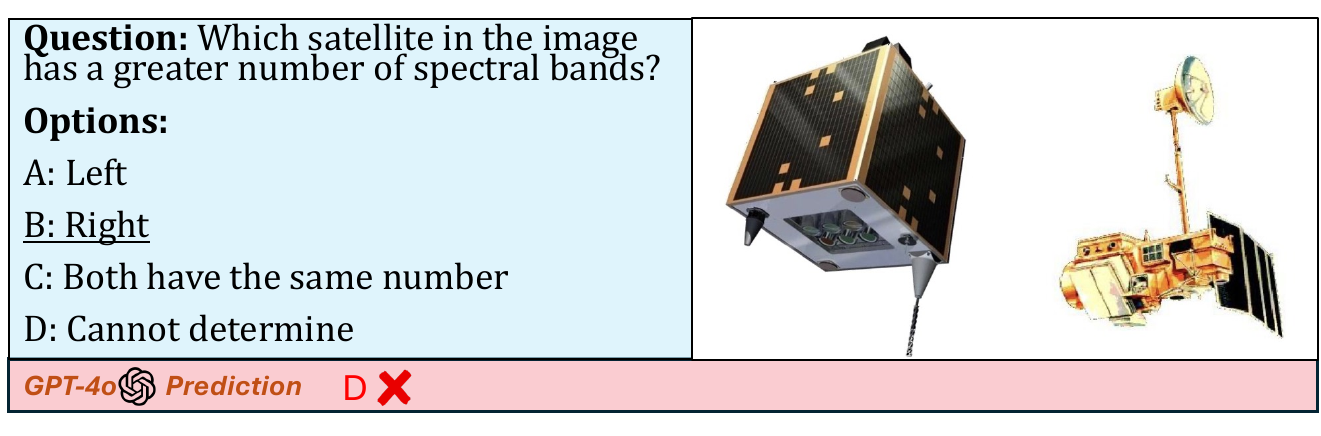}
    \vspace{-20pt}
    \caption{A sample error case in remote sensing platforms. Subfield: Remote sensing satellite identification.}
    \vspace{-10pt}
\end{figure}

\begin{figure}[h]
    \centering
    \includegraphics[width=\linewidth]{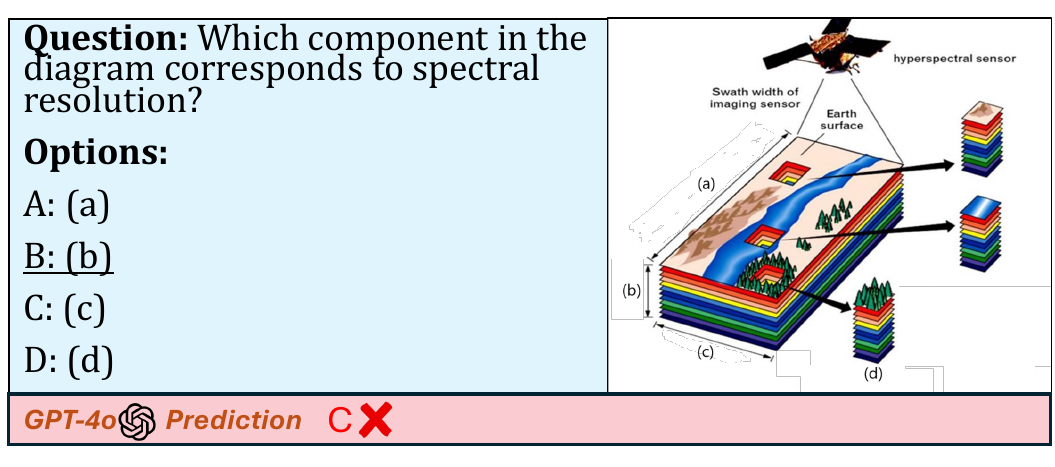}
    \vspace{-20pt}
    \caption{A sample error case in remote sensing principles. Subfield: Resolution type recognition.}
    \label{fig:enter-label}
    \vspace{-10pt}
\end{figure}

\begin{figure}[h]
    \centering
    \includegraphics[width=\linewidth]{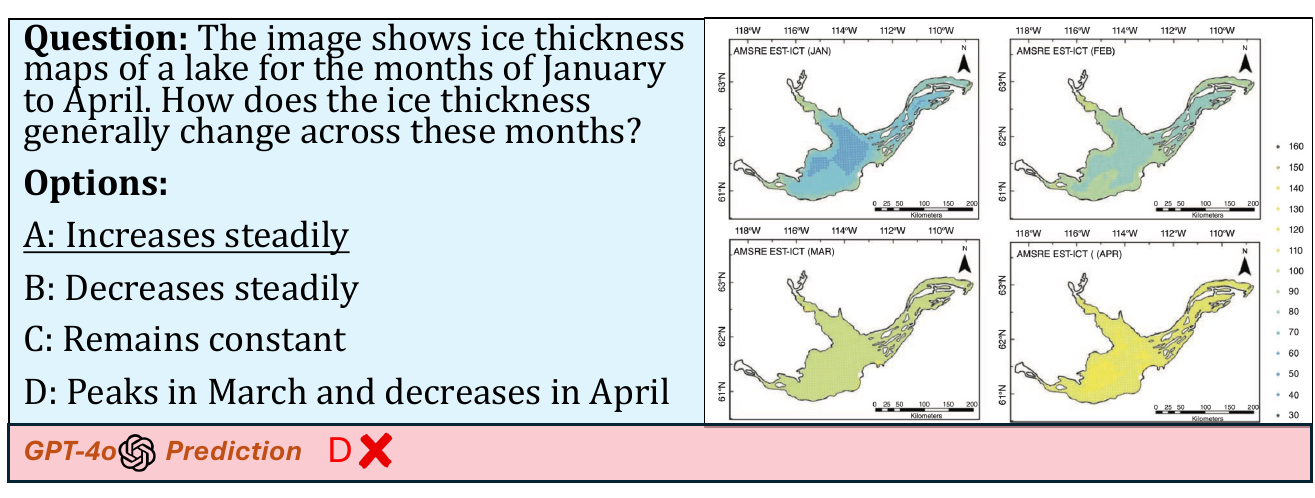}
    \vspace{-20pt}
    \caption{A sample error case in time series analysis. Subfield: Environmental monitoring.}
    \vspace{-5pt}
\end{figure}

\begin{figure}[h]
    \centering
    \includegraphics[width=\linewidth]{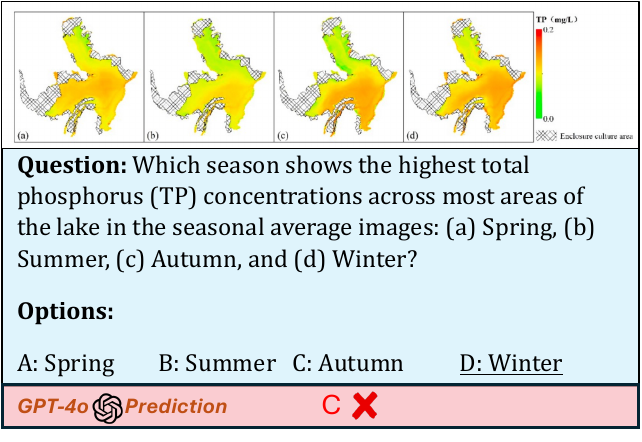}
    \vspace{-20pt}
    \caption{A sample error case in time series analysis. Subfield: Pollution analysis.}
    \vspace{-10pt}
\end{figure}

\begin{figure}[h]
    \centering
    \includegraphics[width=\linewidth]{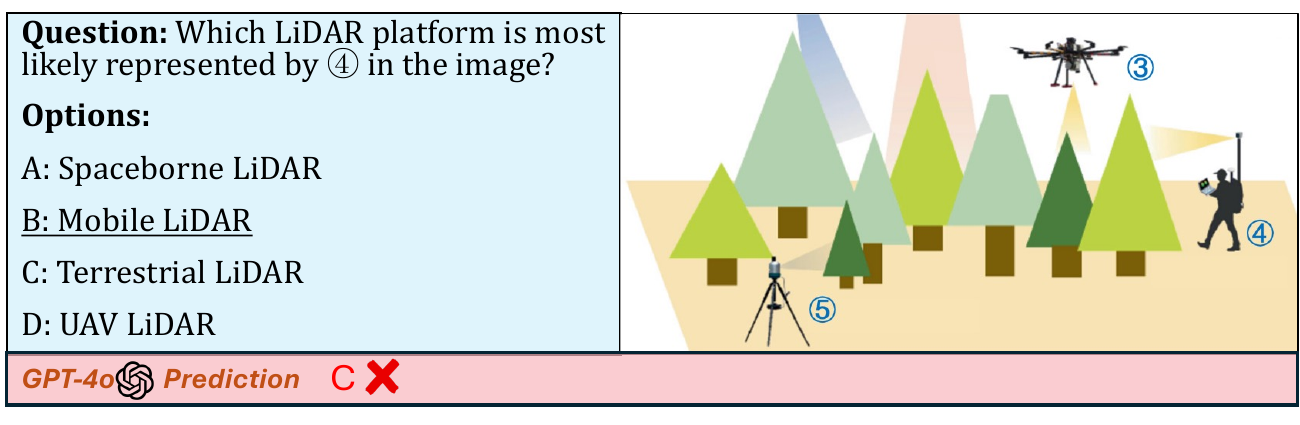}
    \vspace{-20pt}
    \caption{A sample error case in remote sensing principles. Subfield: LiDAR platforms.}
    \vspace{-10pt}
\end{figure}

\begin{figure}[h]
    \centering
    \includegraphics[width=\linewidth]{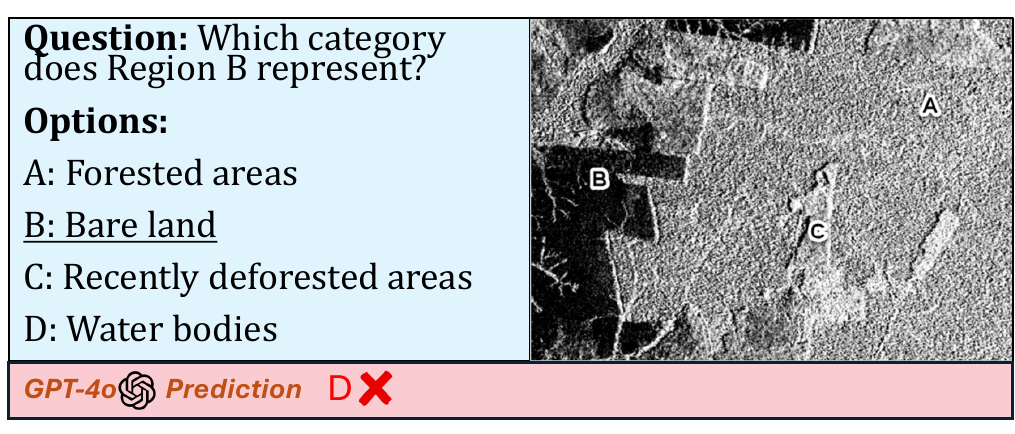}
    \vspace{-20pt}
    \caption{A sample error case in remote sensing perception. Subfield: SAR imagery perception.}
    \vspace{-10pt}
\end{figure}

\begin{figure}[h]
    \centering
    \includegraphics[width=\linewidth]{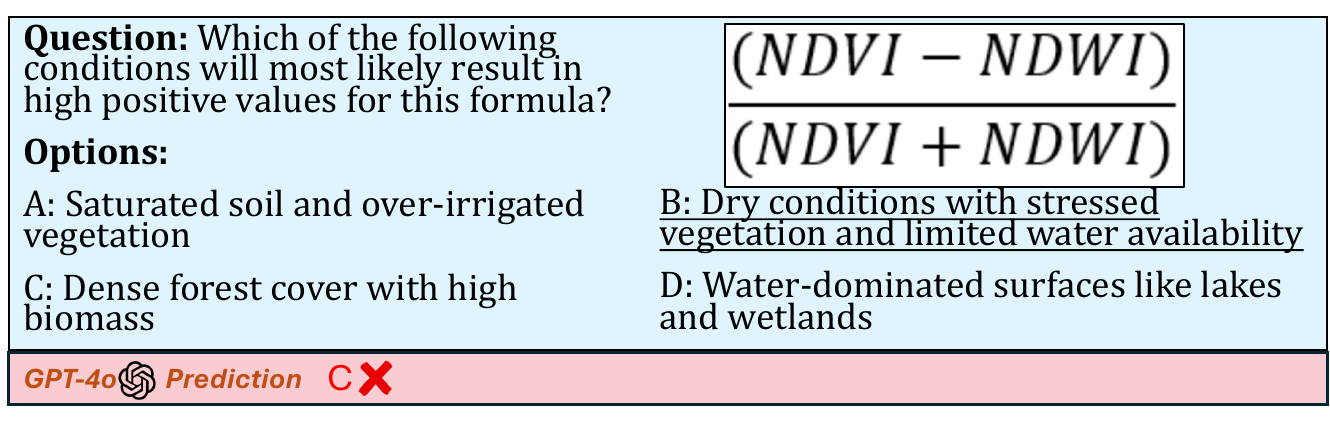}
    \vspace{-20pt}
    \caption{A sample error case in hyperspectral theoretical understanding. Subfield: Spectral index understanding.}
\end{figure}

\begin{figure}[h]
    \centering
    \includegraphics[width=\linewidth]{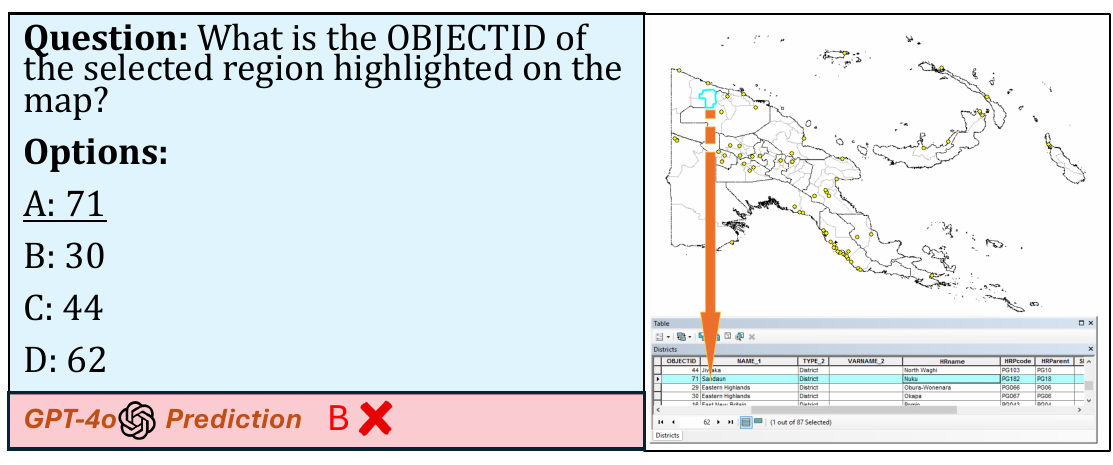}
    \vspace{-5pt}
    \caption{A sample error case in GIS interpretation. Subfield: GIS map recognition.}
    \vspace{-5pt}
\end{figure}

\begin{figure}[h]
    \centering
    \includegraphics[width=\linewidth]{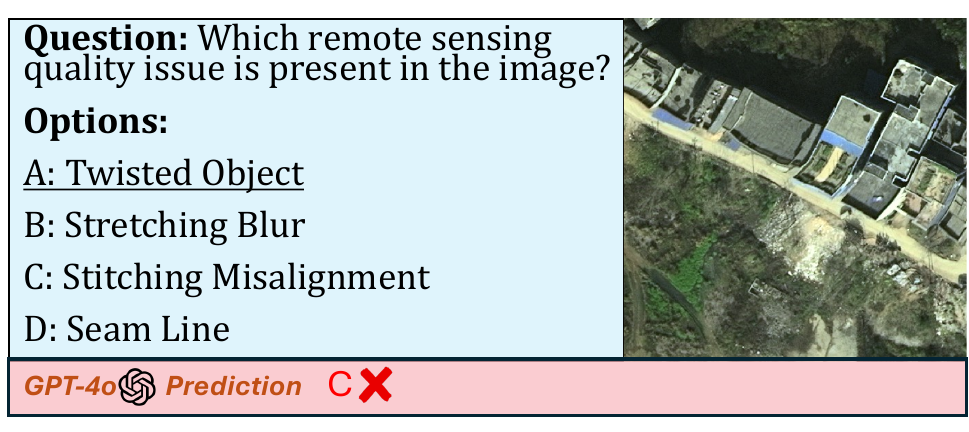}
    \vspace{-5pt}
    \caption{A sample error case in data quality examination. Subfield: Image quality inspection.}
    \vspace{-5pt}
\end{figure}

\begin{figure}[h]
    \centering
    \includegraphics[width=\linewidth]{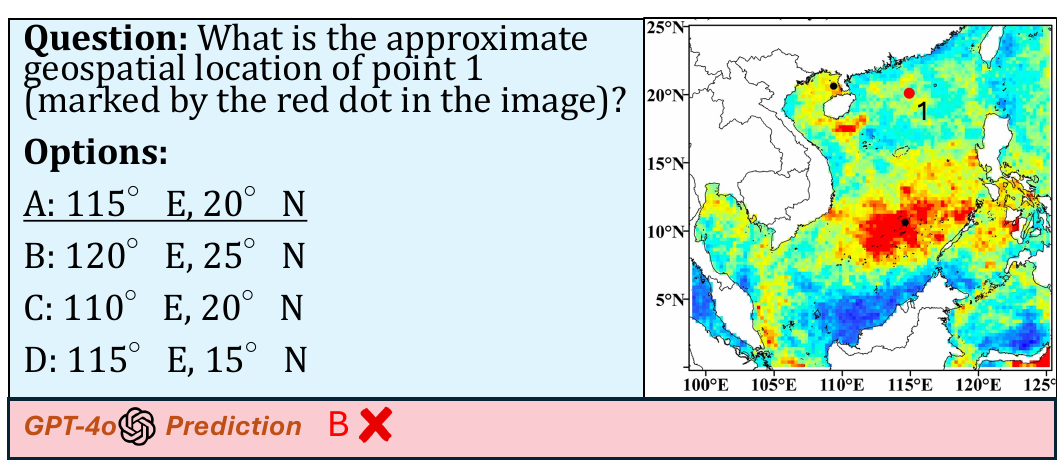}
    \vspace{-5pt}
    \caption{A sample error case in spatial relation analysis. Subfield: Geospatial location identification.}
    \vspace{-5pt}
\end{figure}

\begin{figure}[h]
    \centering
    \includegraphics[width=\linewidth]{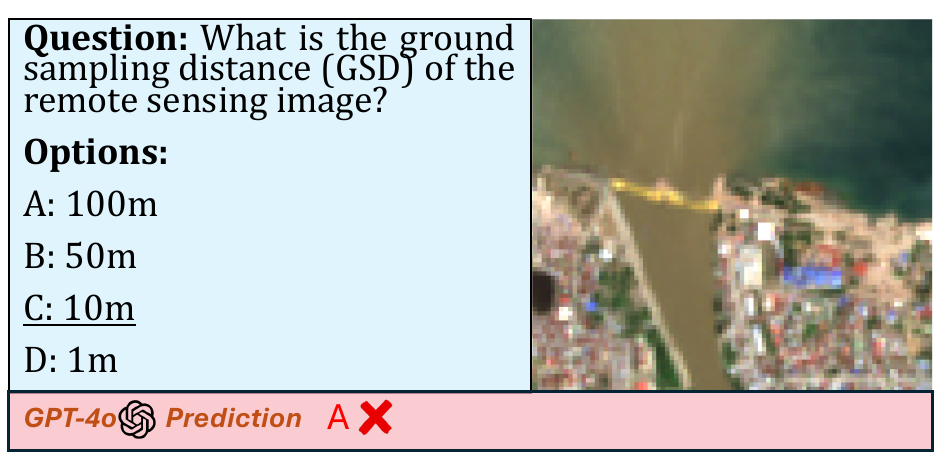}
    \vspace{-5pt}
    \caption{A sample error case in remote sensing image recognition. Subfield: GSD recognition.}
     \label{fig:15}
    \vspace{-5pt}
\end{figure}


\section{More Descriptions on GeoMMAgent}

\subsection{Toolkit Library}

We present the tools integrated into GeoMMAgent. As shown in Fig. 4 and Section 3 of the manuscript, the toolkit library is organized into four categories: \textit{general toolkit}, \textit{knowledge toolkit}, \textit{perception toolkit}, and \textit{reasoning toolkit}. GeoMMAgent is designed as a fully training free and extensible framework, where new tools can be seamlessly added without any model fine tuning or architectural modification. This plug and play design ensures superior flexibility, adaptability to emerging geospatial tasks, and long term upgradeability as external tools and APIs evolve. Below, we provide detailed descriptions of each toolkit category.

\noindent\textbf{General Toolkit} provides essential preprocessing and postprocessing utilities that ensure proper data formatting, quality control, and task specific output handling. These tools serve as the foundation for downstream specialized agents and enable robust execution across diverse geospatial tasks.
\begin{itemize}
    \item \textit{Format conversion}: Converts between different data formats to ensure compatibility across data sources and models.
    \item \textit{Patch tiling and merging}: Divides large images into manageable tiles for efficient processing and later aggregates individual predictions into a unified output.
    \item \textit{Filtering}: Applies smoothing, denoising, or sharpening operations to improve data quality and enhance downstream perception performance.
    \item \textit{Cropping}: Extracts user specified or automatically determined regions of interest from large scale imagery. This reduces irrelevant spatial context, lowers computational cost, and ensures that subsequent modules focus on the most relevant areas.
    \item \textit{Scaling}: Resizes imagery to the resolution or aspect ratio required by downstream modules. It supports both upsampling and downsampling and maintains consistent data formats when combining multi sensor or multi scale inputs.
    \item \textit{Super resolution}: Improves the spatial resolution of remote sensing images using learning based or model based algorithms. This enhances visibility of fine grained structures.
    \item \textit{Area counting}: Measures the total surface area of a specific region or object class after segmentation or thresholding.
    \item \textit{Box counting}: Counts the number of detected bounding boxes for objects of interest. 
\end{itemize}

\begin{table*}[!t]
    \centering
    \renewcommand{\arraystretch}{1.5}
    \setlength{\tabcolsep}{5pt}
    \begin{tabular}{p{3.5cm}|p{12cm}}
    \hline
    \textbf{Agent Role} & \textbf{System Prompt Content} \\
    \hline
    \textbf{Coordinate Agent} & You are an Intelligent Orchestration Expert in the field of Remote Sensing that analyzes images and queries, creating execution plans through the coordination of multiple specialized agents. You provide a detailed description of the inputs and deliver professional task decomposition informed by a meticulous review of the agents' and toolkits' documentation. \\
    \hline
    \textbf{Perception Agent} & You are a Specialized Perception Expert in the field of Remote Sensing, responsible for extracting reliable visual evidence from multi-sensor imagery. You perform scene classification, object detection, and semantic segmentation, and provide calibrated predictions for downstream reasoning. \\
    \hline
    \textbf{Knowledge Agent} & You are a Geospatial Knowledge Retrieval Expert in the field of Remote Sensing, specialized in querying external knowledge bases and web resources. You retrieve, filter, and summarize factual information about remote sensing, geophysics, and geographic entities. \\
    \hline
    \textbf{Reasoning Agent} & You are an Expert Reasoning Agent in the field of Remote Sensing, specialized in multimodal geospatial reasoning. You integrate visual features, retrieved knowledge, and task context to perform step-by-step analysis and produce logically consistent answers. \\
    \hline
    \textbf{Self-Evaluation Agent} & You are a Professional Assessment Expert in the field of Remote Sensing, specialized in evaluating the correctness of image analysis results. You assess logic, consistency, completeness. \\
    \hline
    \end{tabular}
    \caption{Representative system prompts assigned to different agents within the GeoMMAgent framework.}
    \label{tab:agent_prompts}
\end{table*}

\noindent\textbf{Knowledge Toolkit} enables GeoMMAgent to retrieve specialized information from external sources and augment its internal knowledge. This capability is crucial for tasks that involve dynamic environments, region specific context, or domain specific terminology or data  that general purpose pre training does not fully capture. By integrating diverse knowledge bases, the agent can access factual, up to date, and geographically grounded information to support reliable geospatial reasoning.
\begin{itemize}
    \item \textit{Google API}: Provides access to open domain information and real time web content. It is used to verify geographical facts, gather updates on recent environmental events, and supplement internal knowledge with external data. This strengthens the agent’s ability to handle dynamic scenarios.
    \item \textit{Wikimedia API}: Offers structured encyclopedic knowledge about geospatial entities, geological terms, and landmarks. It helps the agent interpret technical vocabulary, understand the characteristics of landforms, and retrieve detailed descriptions of points of interest. 
    \item \textit{GME}~\cite{zhang2025gmeimprovinguniversalmultimodal}: Acts as a semantic alignment engine for queries that combine image inputs with text descriptions. It maps both the fused query and the retrieval candidates into a unified embedding space to evaluate semantic similarity. Unlike traditional text based or image based retrieval methods, this multimodal module identifies evidence that matches the joint visual and textual context, reduces hallucination, and improves the reliability of image grounded reasoning.
\end{itemize}

\noindent\textbf{Perception toolkit} includes expert remote sensing models at the scene level, object level, and pixel level. These models support core geospatial perception tasks and are robust to variations in image resolution. They provide accurate and interpretable outputs that anchor the agent’s reasoning in reliable visual evidence.

\begin{itemize}
    \item \textit{Scene classification model}: We train a Yolo11 based classifier \cite{khanam2024yolov11} with backbone CSPNet on the Million-AID \cite{long2021creating} dataset. The model recognizes 51 scene categories and land cover types, covering the major classes commonly used in remote sensing scene understanding. The toolkit outputs top five predictions with confidence scores to support precise interpretation of scene semantics.
    
    \item \textit{Detection model}: We deploy a pre trained Yolo11 detector \cite{khanam2024yolov11} with backbone CSPNet trained on the DOTA-v2 \cite{Xia_2018_CVPR} dataset. It employs oriented bounding boxes to detect and localize diverse geospatial objects such as aircraft, vehicles, and buildings. The toolkit outputs object counts, spatial distributions, and detection reports that include class labels and confidence values.
    \item \textit{Segmentation model}: We train a DeepLabv3 plus model with Xception backbone \cite{chen2018encoder} on the LoveDA dataset \cite{wang2021loveda} to perform semantic segmentation and pixel level classification of remote sensing imagery. The model delineates land cover types, urban structures, and natural features with precise boundaries. The toolkit provides segmentation masks with per pixel class labels that support area measurement and spatial analysis of heterogeneous landscapes.
\end{itemize}

\noindent\textbf{Reasoning toolkit} includes advanced multimodal language models designed for complex logical inference, spatial and temporal understanding, and knowledge integration. These models operate on the outputs of previous agents and generate reliable final answers for geospatial decision making.  

\begin{itemize}
\item \textit{Reasoning Agent}: We employ Qwen VL Max \cite{Qwen-VL}, which combines high quality visual understanding with strong textual reasoning. It integrates perception outputs, retrieved knowledge, the original image, and the question context to perform multi step inference. The agent conducts semantic matching, consistency checking, option filtering, and final answer generation for tasks that require advanced logical or spatial reasoning.
\end{itemize}
The framework is model agnostic, and other multimodal reasoning models can be incorporated without additional training, maintaining full compatibility with the overall agent pipeline.

\subsection{Agent and Tool Prompts}

We summarize the system prompts governing each agent in Table~\ref{tab:agent_prompts}. Furthermore, Table~\ref{tab:toolkit_prompts} details the functional descriptions of the available tools; these descriptions serve as a reference for the Coordinator Agent to assess tool capabilities and facilitate precise tool invocation. Collectively, these specifications define role-specific behaviors and establish a unified guideline for coordinated multi-agent execution.

\begin{table*}[!t]
    \centering
    \small
    \renewcommand{\arraystretch}{1.3}
    \setlength{\tabcolsep}{8pt}
    \begin{tabular}{p{4cm}|p{12.5cm}}
    \hline
    \textbf{Tool} & \textbf{System Prompt Content} \\
    \hline
    \multicolumn{2}{l}{\textit{General Toolkit}} \\
    \hline
    Format conversion & Use this tool to convert inputs between different image and geospatial data formats so that downstream models and tools can directly consume the data. \\
    \hline
    Patch tiling and merging & Use this tool to split large images into tiles for efficient processing and then merge tile-level predictions back into a spatially consistent full-scene result. \\
    \hline
    Filtering & Use this tool to denoise, smooth, or sharpen imagery in order to improve data quality before perception or reasoning. \\
    \hline
    Cropping & Use this tool to crop user-specified or automatically selected regions of interest, removing irrelevant areas and reducing computational cost. \\
    \hline
    Scaling & Use this tool to resize imagery to the resolution or aspect ratio required by subsequent models, supporting both upsampling and downsampling. \\
    \hline
    Super resolution & Use this tool to enhance the spatial resolution of remote sensing imagery and reveal fine-grained structures that are important for detailed analysis. \\
    \hline
    Area counting & Use this tool to compute the surface area of a given region or semantic class based on segmentation or thresholding results. \\
    \hline
    Box counting & Use this tool to count detected objects from bounding-box outputs and summarize object statistics by category. \\
    \hline
    \multicolumn{2}{l}{\textit{Knowledge Toolkit}} \\
    \hline
    Google API & Use this tool to query open-domain web information related to a geospatial question, verify geographic facts, and obtain up-to-date context about environmental events. \\
    \hline
    Wikimedia API & Use this tool to retrieve structured encyclopedic knowledge about places, landforms, and technical terminology in remote sensing and geoscience. \\
    \hline
    GME & Use this tool to perform multimodal semantic retrieval: given an image–text query, rank candidate documents or patches by similarity in a unified embedding space and return the most relevant evidence. \\
    \hline
    \multicolumn{2}{l}{\textit{Perception Toolkit}} \\
    \hline
    Scene classification model & Use this tool when a scene-level category is needed: it classifies remote sensing images into 51 scene and land-cover types, including Dry land, Greenhouse, Paddy field, Terraced field, Meadow, Forest, Orchard, Commercial area, Storage tank, Wastewater tank, Works, Oil field, Mine, Quarry, Solar, Wind, Substation, Swimming pool, Church, Cemetery, Basketball court, Tennis court, Baseball field, Ground track field, Golf course, Stadium, Detached house, Apartment, Mobile home park, Apron, Helipad, Runway, Road, Viaduct, Bridge, Intersection, Parking lot, Roundabout, Pier, Railway, Train station, Rock land, Bare land, Ice land, Island, Desert, Sparse shrub land, Lake, River, Beach, and Dam, and returns top-\(k\) labels with confidence scores. \\
    \hline
    Detection model & Use this tool when object instances are required: it detects and localizes oriented objects, including Plane, Ship, Storage Tank, Baseball Diamond, Tennis Court, Basketball Court, Ground Track Field, Harbor, Bridge, Large Vehicle, Small Vehicle, Helicopter, Roundabout, Soccer Ball Field, and Swimming Pool, and outputs bounding boxes, categories, and confidence scores. \\
    \hline
    Segmentation model & Use this tool when pixel-wise masks are needed: it produces semantic segmentation maps that delineate land-cover types and structures, enabling area measurement and spatial pattern analysis. \\
    \hline
    \multicolumn{2}{l}{\textit{Reasoning Toolkit}} \\
    \hline
    Spatial-Temporal Analysis & Use this tool to analyze multi-temporal or multi-sensor remote sensing data, characterize spatial–temporal patterns, and reason about changes, trends, and dynamic processes across time. \\
    \hline
    Multiple Choice Matching & Use this tool to align free-form model answers with discrete options in multiple-choice questions, selecting the option that is most semantically consistent with the reasoning outcome. \\
    \hline
    \end{tabular}
    \caption{Representative system prompts assigned to different tools within the GeoMMAgent framework.}
    \label{tab:toolkit_prompts}
\end{table*}

\subsection{Example Cases}

We provide example cases to demonstrate how GeoMMAgent operates as a professional expert in geoscience and remote sensing. Each case is presented in a step by step manner to illustrate how the agent coordinates its toolkits, integrates multimodal evidence, and performs reliable geospatial reasoning.

\subsubsection{Example Case \#1}

\begin{figure}[h]
    \centering
    \includegraphics[width=\linewidth]{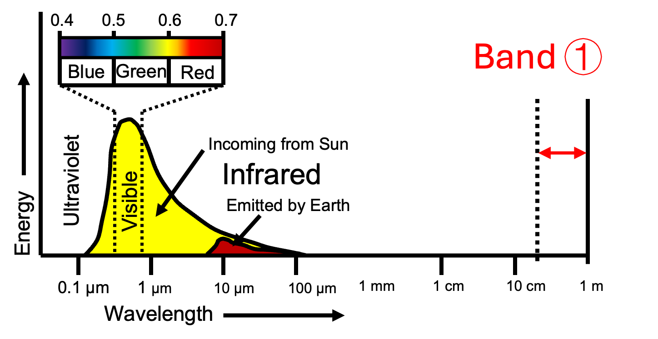}
    \caption{The multimodal query for our case study. The question is: "Which of the following statements about Band \textcircled{1} is accurate?" The options are: (A) It is suitable for thermal remote sensing, (B) It is used in passive remote sensing, (C) It operates only during the day and is affected by weather conditions, and (D) It can penetrate vegetation to detect subsurface targets. The answer is D.}
    \label{fig:question1}
\end{figure}

This example showcases the GeoMMAgent's multi-phase reasoning process for analyzing a spectral band identification question (Figure~\ref{fig:question1}).

\noindent\textbf{Phase 1: Task Specification and Decomposition.} The agent analyzes the query and recognizes it as a multiple-choice question about spectral band properties in remote sensing. The image shows an electromagnetic spectrum diagram with Band \textcircled{1} labeled at a specific wavelength region.
\begin{itemize}
    \item \textbf{Input:} User query and image pair.
    \item \textbf{Action (Task Decomposition):} The agent decomposes the task into: (1) Visually analyze the spectrum diagram to identify the wavelength range and position of Band \textcircled{1}. (2) Retrieve knowledge about the properties of bands in that spectral region. (3) Match the retrieved properties with the given options to determine which statement is accurate.
    \item \textbf{Output:} Task specification: identify Band \textcircled{1}'s spectral characteristics and match with options. Required agents: \texttt{Retrieval Agent} and \texttt{Reasoning Agent}.
\end{itemize}

\noindent\textbf{Phase 2: Initial Task Execution.} The agent executes the plan. The \texttt{Reasoning Agent} analyzes the image and identifies Band \textcircled{1} as located in the microwave region of the spectrum (approximately 1 mm to 1 m wavelength). The \texttt{Knowledge Toolkit} is invoked to search for microwave band properties, but the initial search query ``Band \textcircled{1} properties'' returns limited results. The agent generates an initial answer B based on partial reasoning that microwave sensing can be passive.
\begin{itemize}
    \item \textbf{Input:} The task specification, query, and spectral diagram image.
    \item \textbf{Execution Log:} Visual analysis: Band \textcircled{1} identified in microwave region. Knowledge retrieval: limited results. Reasoning: partial match with option B.
    \item \textbf{Output:} Initial answer B: ``It is used in passive remote sensing.''
\end{itemize}

\noindent\textbf{Phase 3: Self-Evaluation and Error Analysis.} The self-evaluation agent reviews the answer and execution log. It identifies that while option B is partially correct (microwave can be passive), the answer lacks sufficient justification and does not fully leverage the visual information. The evaluation notes that the search query was too generic and did not retrieve specific properties that would distinguish between options.
\begin{itemize}
    \item \textbf{Input:} The image-query pair, all four options (A, B, C, D), the initial answer (B), and the reasoning trace log from Phase 2.
    \item \textbf{Action (Error Analysis):} The evaluation log states: ``Low confidence. Answer B is partially correct but lacks strong evidence. The knowledge retrieval was insufficiently specific. Need to refine search strategy to retrieve distinguishing properties of microwave bands, particularly their penetration capabilities.''
    \item \textbf{Output:} Failure status with specific error analysis: insufficient knowledge retrieval depth.
\end{itemize}

\noindent\textbf{Phase 4: Re-execution with Revised Plan.} Based on the evaluation feedback, the agent revises its knowledge retrieval strategy. It searches for more specific queries: ``microwave band penetration vegetation subsurface'' and ``microwave remote sensing properties.'' The refined search successfully retrieves the key property: ``Microwave bands can penetrate vegetation and detect subsurface features.'' The \texttt{Reasoning Agent} then synthesizes this knowledge with the visual confirmation that Band \textcircled{1} is indeed in the microwave region, leading to a confident match with option D.
\begin{itemize}
    \item \textbf{Input:} The revised plan, error analysis, original query, and image.
    \item \textbf{Action (Revised Execution):} Refined knowledge retrieval with specific queries about microwave penetration properties. Visual confirmation of Band \textcircled{1}'s position. Synthesis of knowledge and visual evidence.
    \item \textbf{Output:} Correct answer D: ``It can penetrate vegetation to detect subsurface targets.''
\end{itemize}

\begin{figure}[t]
    \centering
    \includegraphics[width=\linewidth]{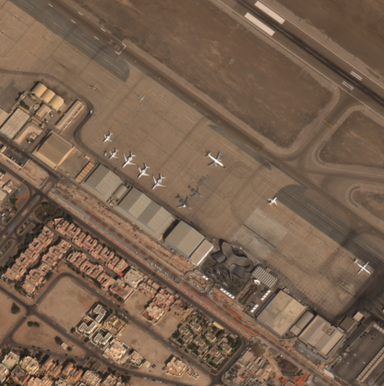}
    \caption{The multimodal query for aircraft counting. The question is: "How many aircraft are there in the image?" The options are: (A) 13, (B) 10, (C) 12, (D) 9. The answer is C.}
    \label{fig:question35}
\end{figure}

\noindent\textbf{Phase 5: Final Self-Evaluation.} The agent performs a final evaluation. The answer D is confirmed as correct: it is supported by (1) visual evidence showing Band \textcircled{1} in the microwave region, (2) retrieved knowledge about microwave penetration properties, and (3) logical matching with option D's description. The evaluation yields high confidence.
\begin{itemize}
    \item \textbf{Input:} The final answer (D) and the complete execution log from Phase 4.
    \item \textbf{Output:} Success status with high confidence, confirming the answer is well-supported by both visual analysis and domain knowledge.
\end{itemize}

\subsection{Example Case \#2}

This case illustrates how GeoMMAgent identify and count specific objects.

\noindent\textbf{Phase 1: Task Specification and Decomposition.} The agent analyzes the query and recognizes it as an object counting task in remote sensing imagery. The image contains multiple aircraft that need to be detected and counted.
\begin{itemize}
    \item \textbf{Input:} User query and image pair.
    \item \textbf{Action (Task Decomposition):} The agent decomposes the task into: (1) Use the \texttt{Detection Toolkit} to detect all aircraft in the image. (2) Count the number of detected aircraft. (3) Match the count with the given options to determine the correct answer.
    \item \textbf{Output:} Task specification: detect and count aircraft in the image. Required agents: \texttt{Detection Agent}.
\end{itemize}

\noindent\textbf{Phase 2: Task Execution.} The agent executes the plan. The \texttt{Detection Toolkit} processes the image and detects aircraft using oriented bounding boxes. The detection successfully identifies 12 aircraft in the image. The count is then matched with the given options, leading to answer C.
\begin{itemize}
    \item \textbf{Input:} The task specification, query, and image.
    \item \textbf{Execution Log:} Detection: 12 aircraft detected with bounding boxes. Count: 12. Match with option C.
    \item \textbf{Output:} Answer C: 12 aircraft.
\end{itemize}

\noindent\textbf{Phase 3: Self-Evaluation.} The agent performs a final evaluation. The answer C is confirmed as correct: it is supported by (1) detection results showing 12 aircraft with bounding boxes, (2) accurate counting of detected objects, and (3) logical matching with option C. The evaluation yields high confidence.
\begin{itemize}
    \item \textbf{Input:} The final answer (C) and the complete execution log from Phase 2.
    \item \textbf{Output:} Success status with high confidence, confirming the count is accurate and well-verified.
\end{itemize}

\subsection{Ablation Study}

To systematically evaluate the contribution of each component in GeoMMAgent, we perform a component-wise ablation study. The results are summarized in Table \ref{tab:ablation}. Overall, all components contribute positively to the final performance. In particular, removing the reasoning module leads to the most significant degradation, highlighting its critical role in the framework. The knowledge and perception modules also provide consistent improvements, while self-evaluation contributes to further refinement, albeit with a comparatively smaller impact.

\begin{table}[t]
    \centering
    \setlength{\tabcolsep}{10pt} 
    \begin{tabular}{l|cc}
    \hline
         & val & test \\
    \hline
       w/o Knowledge  & 83.8 & 87.4 \\
       w/o Perception & 83.8 & 80.3 \\
       w/o Reasoning & 59.5 & 67.3 \\
       w/o Self-evaluation & 81.1 & 80.1 \\
       full GeoMMAgent & 86.5 & 88.4 \\
    \hline
    \end{tabular}
    \caption{Component-wise ablation study of GeoMMAgent on the val and test sets.}
    \label{tab:ablation}
\end{table}

\section{Limitation}
Like any benchmark, GeoMMBench has limitations despite its comprehensive design. The manual curation process may introduce selection biases, and the chosen knowledge points, while diverse, cannot fully represent the complete breadth and depth required for evaluating an Expert AGI in geoscience and remote sensing. Even so, we argue that strong performance on GeoMMBench remains a necessary criterion for an Expert AGI, since it reflects broad domain knowledge, deep subject understanding, and expert level reasoning capabilities.

In terms of system design, the tools integrated into GeoMMAgent are primarily targeted toward the tasks represented in GeoMMBench. Although the toolkit covers many important functionalities, it cannot fully encompass the entire range of tasks present in the wider geoscience and remote sensing community. Nonetheless, GeoMMAgent is built as a fully training free and extensible framework in which new tools can be added seamlessly without model fine tuning or architectural modification. This plug and play design enables high flexibility, adaptability to emerging geospatial applications, and long term upgradeability as external tools, APIs, and models evolve. We plan to incorporate more tools in the future to support an even broader spectrum of geospatial tasks.


\end{document}